\documentclass{article}

\usepackage{arxiv}

\usepackage[utf8]{inputenc} 
\usepackage[T1]{fontenc}    
\usepackage{hyperref}       
\usepackage{url}            
\usepackage{booktabs}       
\usepackage{amsfonts}       
\usepackage{nicefrac}       
\usepackage{microtype}      
\usepackage{lipsum}		
\usepackage{graphicx}
\usepackage{natbib}
\usepackage{doi}
\usepackage{float}
\usepackage{enumerate}

\usepackage{algorithm}
\usepackage{algorithmicx}
\usepackage{algpseudocode}
\usepackage{amsmath}
\usepackage{ntheorem}
\usepackage{balance} 
\usepackage{bm}
\usepackage{booktabs}
\usepackage{tabularx}
\usepackage{diagbox}
\usepackage{multirow}

\usepackage{hyperref}
\usepackage{cleveref}
\usepackage{graphicx}
\usepackage[labelformat=simple]{subcaption} 
\usepackage[many]{tcolorbox}
 \usepackage{tikz}
\usepackage{mathtools}

\newtheorem{theorem}{Theorem}
\newtheorem{definition}{Definition}

\newtheorem*{assumption}{Assumption}

\title{ Homotopy Based Reinforcement Learning with Maximum Entropy for Autonomous Air Combat }

\author{ 
{\includegraphics[scale=0.06]{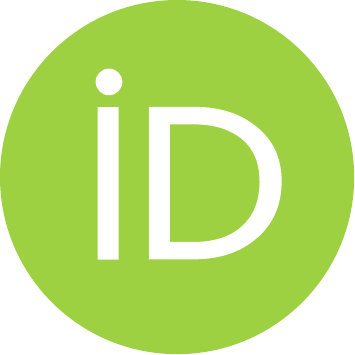}
\hspace{1mm}Yiwen ZHU}\\
	School of Aeronautics and Astronautics\\
	Zhejiang University\\
	Hangzhou 310027, Zhejiang, China \\
	\texttt{113712@zju.edu.cn} \\
	\And
	{\includegraphics[scale=0.06]{orcid.pdf}
	\hspace{1mm}Zhou FANG*}\\
	School of Aeronautics and Astronautics\\
	Zhejiang University\\
	Hangzhou 310027, Zhejiang, China \\
	\texttt{zfang@zju.edu.cn} \\
	\And
	{\includegraphics[scale=0.06]{orcid.pdf}
	\hspace{1mm}Yuan ZHENG}\\
	School of Aeronautics and Astronautics\\
	Zhejiang University\\
	Hangzhou 310027, Zhejiang, China \\
    \And
	{\includegraphics[scale=0.06]{orcid.pdf}
	\hspace{1mm}Wenya WEI}\\
	School of Aeronautics and Astronautics\\
	Zhejiang University\\
	Hangzhou 310027, Zhejiang, China \\
}

\renewcommand{\shorttitle}{\textit{arXiv} Template}

\hypersetup{
pdftitle={A template for the arxiv style},
pdfsubject={q-bio.NC, q-bio.QM},
pdfauthor={David S.~Hippocampus, Elias D.~Striatum},
pdfkeywords={First keyword, Second keyword, More},
}

\begin{document}
\maketitle

\begin{abstract}

The Intelligent decision of the unmanned combat aerial vehicle (UCAV) has long been a challenging problem.
The conventional search method can hardly satisfy the real-time demand during high dynamics air combat scenarios.
The reinforcement learning (RL) method can significantly shorten the decision time via using neural networks.
 However, the sparse reward problem limits its convergence speed and the artificial prior experience reward can easily deviate its optimal convergent direction of the original task, which raises great difficulties for the RL air combat application.
 In this paper, we propose a homotopy-based soft actor-critic method (HSAC) which focuses on addressing these problems via following the homotopy path between the original task with sparse reward and the auxiliary task with artificial prior experience reward.
 The convergence and the feasibility of this method are also proved in this paper.
To confirm our method feasibly, we construct a detailed 3D air combat simulation environment for the RL-based methods training firstly, and we implement our method in both the attack horizontal flight UCAV task and the self-play confrontation task.
 Experimental results show that our method performs better than the methods only utilizing the sparse reward or the artificial prior experience reward.
 The agent trained by our method can reach more than $98.3\%$ win rate in the attack horizontal flight UCAV task and average $67.4\%$ win rate when confronted with the agents trained by the other two methods.
\end{abstract}


\section{Introduction}\label{SE:Introduction}

%


Unmanned combat aerial vehicle (UCAV) is an essential component in future air combat. For now, the UCAV has the ability to accomplish some regular tasks with the assistance of a ground station, including reconnaissance and detection, and tracking. In the future air combat environment, on account of its lower cost of manpower and fewer constraints on dynamic, UCAV will also be the main force of air confrontation. 


Air combat tactics decision-making is one of the most important techniques in the autonomous process, and it has become especially crucial when UCAVs were integrated into the dynamic combat environments.
If the decision-making problem can be solved by UCAVs themselves, the communication restrictions with ground stations will be no longer exist, which can truly achieve the autonomy of UCAVs.


The idea of autonomous air combat for the UCAV has been explored by a number of scholars. Xu G and Park H\cite{xu2017application}\cite{park2016differential} used the differential game method in air combat problems.
In the paper published by Virtanen K and Lin Z\cite{virtanen2006modeling}\cite{lin2007sequential}, the influence diagram method had been used to model the maneuvering decisions of pilots in the one-to-one air combat scenario. Nonlinear programming\cite{horie2006optimal} and model prediction control method \cite{ortiz2021comparative} are also used in modeling the air combat problem. In addition, methods such as genetic algorithm\cite{smith2000classifier} and bayesian inference \cite{changqiang2018autonomous} also had been used in autonomous decision-making in air combat scenarios. Although these traditional methods can establish the relationship of different elements in air combat scenarios, they can hardly figure out the problem in a complex environment because of the limitation of real-time calculation.


Recently, artificial intelligent methods are also highlighted in the air combat decision-making problem for the highly real-time of artificial intelligent methods. Methods based on artificial intelligence mainly include the expert system, supervised learning, and RL-based methods, etc.
The expert system method, which is composed of empirical policies, is used in decision-making problem\cite{shenyu1999research}\cite{zhao2008application}\cite{bechtel1992air}.
By training a neural network, the supervised learning method learns the policy of UCAVs from a large dataset of real air combats, solving the decision-making problem
\cite{rodin1992maneuver}\cite{schvaneveldt1992neural}\cite{teng2012self}.


Compared with the expert system and the supervised learning approach, RL-based methods have many advantages. 
 Unlike the expert system method and the supervised learning approach, RL-based methods do not require the strict strategy design and huge dataset for the policy updating, it can acquire the dataset directly by interacting with the external environment
 \cite{sutton2018reinforcement}. 
 Therefore, a great number of scholars, researching air combat decision-making, pay more attention to RL-based methods. In the beginning, Jonathan and James\cite{mcgrew2010air} used the approximate dynamic programming method to establish the maneuver decision model in a 2D 1v1 air combat environment, and
 this method had been tested in the real world with micro-UCAV.
This idea has been confirmed that via RL-based methods the autonomous maneuver decision can be realized in air combat. In 2018, Xiaoteng Ma \cite{ma2018air} with his team in Tsinghua University extend the work of Jonathan \cite{mcgrew2010air}. In this paper, different from the work of Jonathan \cite{mcgrew2010air} the speed control actions had been added into the discrete action space and the deep Q-learning (DQN) method had been used here. In 2020, Zhuang Wang \cite{wang2020improving} proposed an alternate freeze game framework to deal with nonstationarity problems, which adopted the league system to solve the problem of variable opponent strategies in air combat. In the work published by Qiming Yang\cite{yang2019maneuver}, a second-order UCAV motion model and one-to-one short-range air combat model in 3D space are established, using DQN and "basic-confrontation" training method to carry out the air combat autonomous maneuver decision model.


RL-based methods have been proven in the above works that can be used in the air combat scenario. However, these works oversimplify the air combat scenario. Numerous works are based on the assumptions that the UCAVs in air combat are moving in a 2D plane or the action space of UCAVs are dispersed as the basic flight maneuvers (BFM). These assumptions are too far from reality, leading to the limited performance of explorable policies and the policy trained by these models can hardly transfer to the actual air combat scenarios.
Another problem for RL-based methods is the sparse reward problem\cite{sutton2018reinforcement}. Especially, when facing high dimensional programming problems with sparse rewards setting such as air combat scenarios, the agent has to discover a long sequence of "correct" actions in order to achieve the sparse reward signal. 
Usually, It is too hard for the agent to discover this sparse reward via random exploration. Therefore, the learning agent has little to no feedback on the quality of its actions.

To deal with this sparse reward problem, a multitude of methods have been developed, such as shaping rewards\cite{ng1999policy}\cite{randlov1998learning}\cite{gu2017deep}, curriculum learning\cite{heess2017emergence}\cite{ghosh2017divide}\cite{forestier2017intrinsically}, learning from demonstrations\cite{ross2011reduction}\cite{vecerik2017leveraging}\cite{kober2011policy}, learning with model guidance \cite{montgomery2016guided}, and inverse RL\cite{ziebart2008maximum} etc.
Unfortunately, all of these methods are rely too much on the artificial prior knowledge of the specific task.
And they always bias the certain converging direction of the policy, potentially suboptimal direction.

To solve the sparse reward problem of RL-based methods, a homotopy-based soft actor-critic (HSAC) algorithm is developed in this paper and the convergence of this method has been proved, which can be suited in any RL method. 
Furthermore, we apply the HSAC method in the air combat scenario with the idea of self-play which realizes finding the desired equilibrium point \cite{lemke1984pathways} with good offensive and defensive characteristics in the policy space.

Generally speaking, the air combat simulation system is difficult to construct, because the degree of simplification of the model is difficult to grasp.
Meanwhile, the sparse reward problem of RL-based methods for the relatively complex environment is the bottleneck for the RL application. 
In allusion to these problems, this paper impresses its contribution by the following points:

1. We model the air combat environment with a new continuity modeling method. This method overcomes the problem of limited performance of the explored policy and the difficulty of transferring the policy to the real air combat scenario, which is caused by the difference between the traditional discrete modeling method, e.g. BFM-based modeling method, and the reality. 


2. Because of the sparse reward problem commonly existing in RL problems, we propose a homotopy-based soft actor-critic method and prove the feasibility and convergence of the method theoretically.
This study provides a new perspective for the exploration of RL-based methods.

3. Combined self-play, HSAC is applied to air combat scenarios. Simulation results show that the agent using HSAC performs better than the agents trained by the methods only utilizing the sparse reward or the artificial prior experience reward (average $99.6\%$ and $67\%$ win rate in two different tasks).


This paper is organized as follows: 
The problem of air combat and the RL-based methods are introduced in Section \ref{SE:preliminary}.
The design of the one-to-one air combat environment for training the agent via RL-based methods is stated in Section ~\ref{SE:Env}.
In Section ~\ref{SE:method}, we propose the HSAC method and give proof of the convergence and feasibility of this method.
The training process and the experimental results of the attack horizontal flight task and the confrontation task are exhibited in Section \ref{SE:Sim}.
Conclusions are stated in Section ~\ref{SE:Conclusion}.

\section{Preliminary}\label{SE:preliminary}

In this section, we will introduce the dynamics model of UAV short-range air combat and the basic principles of RL-based methods especially the soft actor-critic method.

     \subsection{One-To-One Short-Range Air Combat Problem}\label{SE:Air_combat_model}
    The aim of short-range air combat, also called dogfight, is to shoot down the opponent's UCAV while avoiding being shot down by the opponent's UCAV.
    We use superscript $b$ and superscript $r$ to distinguish both sides' UCAVs in air combat scenarios. $\text{UCAV}^b$ means the blue side's UCAV, and $\text{UCAV}^r$ denotes the red side's UCAV.
    The relationship between $\text{UCAV}^b$ and $\text{UCAV}^r$ is adversarial.
    Because the superscripts $b$ and $r$ are defined subjectively, these superscripts are interchangeable.
    For the sake of expression, we just consider the decision-making of $\text{UCAV}^b$ in the following paragraphs.

\begin{figure}[ht]
    	\centering
    	\includegraphics[width = 2.5in]{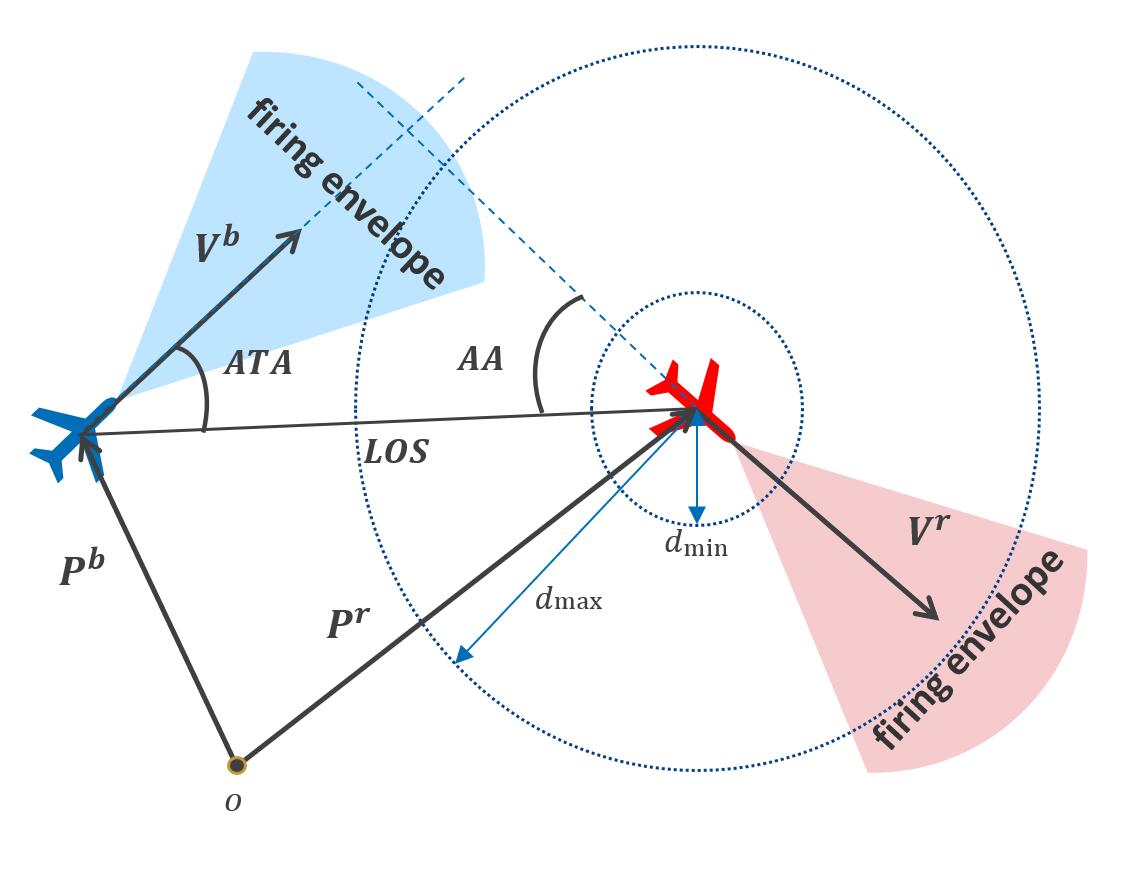}
    	\caption{One-to-one short-range air combat scenario}
    	\label{Fig:1v1AirCombat}
    \end{figure}

    
    The situation of one-to-one short-range air combat is shown in Figure~\ref{Fig:1v1AirCombat}. 
    The concepts, shown in this figure, are used to describe the advantages of $\text{UCAV}^b$ include Aspect Angle (AA), Antenna Train Angle (ATA), Line of Sight (LOS), and Relative Distance.
    If we know the velocity vector of $\text{UCAV}^b$ (${V^b}$) and the position vector of $\text{UCAV}^b$ ($P^b$), the velocity vector of $\text{UCAV}^r$ (${V^r}$) and the position vector of $\text{UCAV}^r$ ($P^r$) can be calculated.

     If $\text{UCAV}^b$ wants to shoot down $\text{UCAV}^r$, the position of $\text{UCAV}^b$ needs to satisfy the features below:
    \begin{enumerate}[(1)]
    \item The relative distance of these two UCAVs between the maximum attack range and the minimum collide range.
    \item $\text{UCAV}^b$ needs to be in an advantageous position.
    In other words, $\text{UCAV}^b$ needs to be in the position where $\text{UCAV}^b$  can easily pursue $\text{UCAV}^r$ and hard to be attacked simultaneously.
    \end{enumerate}
    
    These features can be described by mathematics with inequations, as shown in Eq.~(\ref{Eq:attackEquation})\cite{mcgrew2010air}.
    \begin{align}
    \begin{dcases}
        {d_{\min }} < {d_{LOS}} < {d_{\max }}\\
        \left| {AA^b} \right| < {60^ \circ }\\
        \left| {ATA^b} \right| < {60^ \circ }
    \end{dcases}   
    \label{Eq:attackEquation}
    \end{align}
    If $\text{UCAV}^b$ satisfies all of these constraints in Eq.~(\ref{Eq:attackEquation}), $\text{UCAV}^r$ will be in the firing envelope of $\text{UCAV}^b$ \cite{shaw1985fighter},
    and the position of $\text{UCAV}^b$ is a subset of advantageous position of $\text{UCAV}^b$.
    These principles are the same  for $\text{UCAV}^r$.
    
    Therefore, this paper focuses on teaching the UCAV to reach the advantageous position with less time. 
    And  we can use a two-target differential game model\cite{grimm1991modelling}\cite{blaquiere1969quantitative} to describe the air combat problem, as show in Eq.~(\ref{Eq:problemOfAirCombat}).
        \begin{align}
            \begin{split}
                \mathop {\min }\limits_{{u^b}} \mathop {\max }\limits_{{u^r}} &J({u^b},{u^r})=
                \\&\quad  f({t_f},{\mathbf{x}^b}({t_f}),{\mathbf{x}^r}({t_f}))+ \int\limits_{{t_0}}^{{t_f}} {\mathcal{L}_b(t,{\mathbf{x}^b},{\mathbf{x}^r},{u^b},{u^r})dt}
                \\
                \mathop {\min }\limits_{{u^r}} \mathop {\max }\limits_{{u^b}} &J({u^r},{u^b})= \\
                &\quad f({t_f},{\mathbf{x}^r}({t_f}),{\mathbf{x}^b}({t_f}))+ \int\limits_{{t_0}}^{{t_f}} {\mathcal{L}_r(t,{\mathbf{x}^r},{\mathbf{x}^b},{u^r},{u^b})dt} 
            \end{split} 
            \label{Eq:problemOfAirCombat}
        \end{align}
        where $u^b$ and $u^r$ represent the control laws of each UCAV, also can be called the policy. $\mathbf{x}^b$ and $\mathbf{x}^r$ are the state vectors of these two UCAVs, as shown in Eq.~(\ref{Eq:xstate}).
        \begin{align}
            \label{Eq:xstate}
            \begin{split}
                {\mathbf{x}^b} &= {[{x^b},{y^b},{z^b},{v^b},{\gamma ^b},{\chi ^b}]^ \mathrm{ T }}\\
                {\mathbf{x}^r} &= {[{x^r},{y^r},{z^r},{v^r},{\gamma ^r},{\chi ^r}]^ \mathrm{ T }}
            \end{split}
        \end{align}
        $f(\cdot)$ means the terminal punishment function, $\mathcal{L}_b(\cdot)$ and $\mathcal{L}_r(\cdot)$ denote the loss functions of $\text{UCAV}^b$ and $\text{UCAV}^r$, respectively. 
        As all we can see, it is just an optimization problem that is subject to the dynamic constraints of the UCAV introduced in Section~\ref{SE:UCAV_dynamic_model}.

    \subsection{UCAV dynamics model}\label{SE:UCAV_dynamic_model}
     
\begin{figure}[ht]
    	\centering
    	\includegraphics[width = 2.5in]{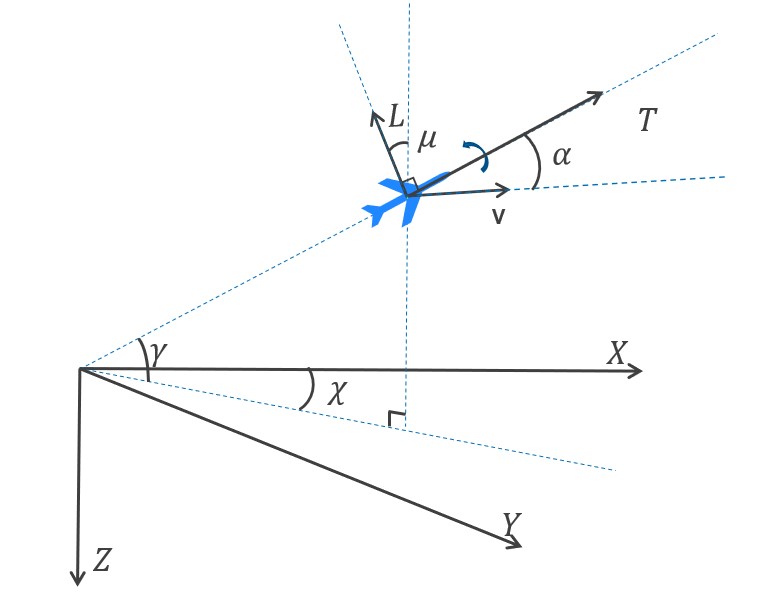}
    	\caption{Three-degree-of-freedom motion model of the UCAV in ground coordinate system}
    	\label{Fig:3-dmotionModel}
    \end{figure}

    UCAV dynamics model is the basis of air combat confrontation.
    This model has been built in the ground coordinate system.
    In this reference system, the equations of motion of the UCAVs can be described concisely, as shown in Eq.~(\ref{Eq:3-dmotionModel}).
    \begin{align}
    	\begin{dcases}
          \dot{x} &= v\cos \gamma \cos \chi  \hfill \\
          \dot{y} &= v\cos \gamma \sin \chi  \hfill \\
          \dot{z} &=  - v\sin \gamma  \hfill \\ 
        \end{dcases}
    	\label{Eq:3-dmotionModel}
    \end{align}
    Where $\dot{x}$, $\dot{y}$, $\dot{z}$ represent the position change rate of the UCAV in  $X$, $Y$, $Z$ axis, respectively. The remaining three state variables in Eq.~(\ref{Eq:3-dmotionModel}) are the flight path angle $\gamma$, the heading angle $\chi$, and the velocity of UCAV $v$. The flight diagram is shown in Figure~\ref{Fig:3-dmotionModel}.
    
    The state variables $v$, $\gamma$, $\chi$ were guided with the control variables: attack angle $\alpha$, throttle setting parameter $\eta$, and the bank angle $\mu$. The point mass model of an UCAV is described by the following system of differential equations in Eq.~(\ref{Eq:pointMassModel})\cite{virtanen2006modeling}.
    \begin{align}
    \begin{dcases}
    \dot v = \frac{1}{m}\{ \eta {T_{\max }}\cos \alpha  - D(\alpha ,v)\}  - g\sin \gamma \\
    \dot \chi  = \frac{1}{{mv\cos \gamma }}\{ \eta {T_{\max }}\sin \alpha  + L(\alpha ,v)\} \sin \mu \\
    \dot \gamma  = \frac{1}{{mv}}\{ \{ \eta {T_{\max }}\sin \alpha  + L(\alpha ,v)\} \cos \mu  - mg\cos \gamma \} 
    \end{dcases}
    \label{Eq:pointMassModel}
    \end{align}
   Where $g$ is the acceleration caused by the gravity, $m$ denotes the mass of the UCAV, and $T_{\max }$ denotes the maximum available thrust force of the engine.
   Meanwhile, $g$, $m$ and $T_{\max }$ are assumed as constants.
    $L( \cdot )$ means the lift force, and $D( \cdot )$ represents the drag force. The equations of these two forces are given in Eq.~(\ref{Eq:liftAndDragForce}):
    \begin{align}
    \begin{dcases}
    L(\alpha ,v) = \frac{1}{2}\rho {v^2}{S_w}{C_L}(\alpha )\\
    D(\alpha ,v) = \frac{1}{2}\rho {v^2}{S_w}{C_D}(\alpha )
    \end{dcases}
    \label{Eq:liftAndDragForce}
    \end{align}
    Where ${C_L}( \cdot )$ and ${C_D}( \cdot )$ are the lift coefficient and drag coefficient, respectively. ${S_w}$ denotes the reference wing area. $\rho$ means the air density, which is assumed as a constant in this work because the range of altitude in air combat scenarios is small. Also the lift coefficient ${C_L}( \cdot )$ and drag coefficient ${C_D}( \cdot )$ are simplified, so the function only related to attack angle $\alpha$, as given in Eq.~(\ref{Eq:liftAndDragCoefficient}).
    \begin{align}
        \begin{dcases}
         {C_L}(\alpha ) = {C_{L0}} + {C_{L\alpha }}\alpha \\
         {C_D}(\alpha ) = {C_{D0}} + BDP \cdot {C_L}{(\alpha )^2}
        \end{dcases}
        \label{Eq:liftAndDragCoefficient}
    \end{align}
    
    $C_{L0}$, $C_{L\alpha }$, $C_{D0}$, $BDP$ in Eq.~(\ref{Eq:liftAndDragCoefficient}) are assumed as constants, which represent the zero-lift coefficient, derivative of lift coefficient with respect to attack angle, zero-drag coefficient, and drag-lift coefficient, respectively. 
    
    The control variables as well as their rates of change are constrained by lower and upper bounds.
    For the performance and the inertia of the UCAV, the control variables' rates of change are constrained.
   For the short-range air combat game, we suppose the throttle setting parameter as constant 1 to maintain the maximum pursue ability.
   The constraints are given in Eq.~(\ref{Eq:controlConstraints}).
    \begin{align}
    \begin{dcases}
    \alpha  \in {\rm{[}}{\alpha _{\min }},{\alpha _{\max }}{\rm{]         }}\\
    \dot \alpha  \in {\rm{[ - }}\Delta \alpha ,\Delta \alpha ]\\
    \mu  \in [ - \pi ,\pi ]\\
    \dot \mu  \in [{\rm{ - }}\Delta \mu ,\Delta \mu ]\\
    \eta  = {\eta _{\max }} = 1{\rm{                  }}\\
    \dot \eta  = 0
    \end{dcases}
    \label{Eq:controlConstraints}
    \end{align}
    
    The load factor $n( \cdot )$ and dynamic pressure $q( \cdot )$ are defined in Eq.~(\ref{Eq:loadFactordef}).
    \begin{align}
    \begin{split}
          n(\alpha ,v) &= \frac{{L(\alpha ,v)}}{{mg}}{\rm{    }}\\
            q(v) &= \frac{1}{2}\rho {v^2}
    \end{split}
    \label{Eq:loadFactordef}
    \end{align}
    
     To avoid overloading of the UCAV, the load factor and dynamic pressure must be limited, as shown in Eq.~(\ref{Eq:dPLimit})\cite{virtanen2006modeling}.
    \begin{align}
    \begin{dcases}
    n(\alpha ,v) - {n_{\max }} \le 0\\
    {h_{\min }} - h \le 0\\
    q(v) - {q_{\max }} \le 0
    \end{dcases}
    \label{Eq:dPLimit}
    \end{align}
    Where air density $\rho$ has already been supposed as a constant, 
    so that the altitude $h$ is uncorrelated to both load factor function $n( \cdot )$ and dynamic pressure function $q( \cdot )$.
    Meanwhile, $h_{\min }$, $n_{\max}$, $q_{\max}$ refer to the minimum altitude, maximum load factor, and maximum dynamic pressure, respectively.
    From the reasons presented above, these variables as well as their constraints are determined by the angle of attack, altitude, and velocity.

  \subsection{Soft actor-critic Method}

    By using Markov Decision Process (MDP), the RL-based methods analyze the problem and let the agents communicate with the environment to acquire the experience automatically so that the policy of the agents can be optimized.
    
    The MDP models the sequential decision-making problems with mathematical formalism. MDP consists of a set of state $S$, a set of action $A$, a transfer function $T$, and a reward function $R$, forming as a tuple $ < S,A,T,R > $. In the time step $t$, 
    given state ${s_t}$, the agent will select an action ${a_t}$, and then the environment will feedback on the next state ${s_{t+1}}$ to the agent according to the transition probability $T({s_{t + 1}}|{s_t},{a_t}) \in [0,1]$.
    After that, the environment return a reward ${r_{t + 1}} = R({s_t},{a_t})$ related to the quality of this transition. The way for the agent to select the action in a state is called a policy $\pi :S \to A$, a mapping from state to the possibility distribution of actions.
    The probability of each action with the given state can be calculated by $\pi (a|s)$. 
    The goal of RL-based methods is to find an optimal policy ${\pi ^*}$ to maximize the expected sum of rewards.
        However, without enough exploration to the state space may be really hard to find the optimal policy $\pi^*$.
        And it is hard for the agent to explore optimally in action space.
        This particular challenge can be addressed with the maximum entropy reinforcement learning methods. 
          Specifically, Soft Actor-Critic (SAC) \cite{haarnoja2018soft}, a model-free and off-policy RL algorithm, is one of the most successful RL algorithms based on the maximum entropy method, which incorporates the policy entropy into the objective function to incentive the exploration of different actions in different states.
          It has become a common baseline algorithm in most of the RL libraries, performing better than most of the other state-of-the-art RL-based methods such as SQL\cite{haarnoja2017reinforcement} and TD3\cite{fujimoto2018addressing} in many environments\cite{haarnoja2018soft} \cite{haarnoja2018soft_2}. 
        The optimal policy $\pi^*$ can be represented in Eq.~(\ref{Eq:softAC}).
        \begin{align}
            {\pi ^*} = \mathop {\arg \max }\limits_\pi  {_{\scriptstyle{s_0} \sim {\rho _0}({s_0}){\rm{ }}{a_t} \sim \pi ({a_t}|{s_t})\hfill\atop
            \scriptstyle{s_{t + 1}} \sim T({s_{t + 1}}|{s_t},{a_t})\hfill}}\left\{ {\sum\limits_{t = 0}^T {{R_{t + 1}}({s_t},{a_t}) + } \alpha \mathrm{H}[\pi(\cdot |{s_t})]} \right\}
            \label{Eq:softAC}
        \end{align}
        Where $H(\cdot|s_t)$ means the entropy of the probability distribution of actions in $s_t$.
        $\rho :S \to \mathbb{R}$ is the probability distribution of the agent’s initial state. $\gamma \in (0,1)$ denotes the discount factor, which can determine if the agent focuses on short-term rewards or long-term rewards.
        $\alpha$ denotes the temperature parameter of SAC, which may considerably affect the convergence of this algorithm \cite{haarnoja2018soft}. 
        And $T$ represents the terminal time step.
        
        Following the Bellman Equation\cite{sutton2018reinforcement}, the soft Q-function can expressed by Eq.~(\ref{Eq:softQ})
        \begin{align}
            Q(s_t,a_t)=r_{t+1}+\gamma \mathbb{E}{_{{s_{t + 1}} \sim T({s_{t + 1}}|{s_t},\pi(s_t))}}[V(s_{t+1
            })]
        \label{Eq:softQ}
        \end{align}
        And the soft value function can be deduced by the soft Q-function, represented as Eq.~(\ref{Eq:softV}).
        \begin{align}
            V(s_t)=\mathbb{E}_{a_t\sim{\pi(s_t)}}[Q(s_t,a_t)-\alpha\log\pi(a_t|s_t)]
            \label{Eq:softV}
        \end{align}
        The parameters of soft Q-function are trained to minimize the temporal difference (TD) error $\delta = r_{t+1}+\gamma v(s_{t+1})-Q(s_t)$. And the parameters of policy $\pi$ are trained to minimize the Kullback–Leibler (KL) divergence\cite{shannon1948mathematical} between the normalized soft Q-function and the probability distribution of the policy, as ${D_{KL}}({\pi _\phi }( \cdot |{s_t})||\frac{{\exp ({Q_\theta }({s_t}, \cdot ))}}{{{Z_\theta }({s_t})}})$, where
        $\theta$ and $\phi$ represent the parameter of critic network and actor network, respectively.
        And
        $\frac{{\exp ({Q_\theta }({s_t}, \cdot ))}}{{{Z_\theta }({s_t})}}$ represents the normalized soft Q-function. 

        So the loss function of soft Q-function is shown in Eq.~(\ref{Eq:JQ}):
        \begin{align}
        \begin{split}
        &{J_Q} =\\
        &{\mathbb{E}_{({s_t},{a_t}) \sim D}}[{\frac{1}{2}}{({Q_\theta }({s_t},{a_t}) - (r({s_t},{a_t})+ \gamma {\mathbb{E}_{{s_{t + 1}} \sim }}_T[{V_{\tilde \theta }}({s_{t + 1}})]))^2}]
        \end{split}
            \label{Eq:JQ}
        \end{align}
        and the cost function of policy can be simplified to Eq.~(\ref{Eq:JPi}):
        \begin{align}
            {J_\pi } = {\mathbb{E}_{{s_t} \sim D}}[{\mathbb{E}_{{a_t} \sim {\pi _\phi }}}[\alpha \log \pi ({a_t}|{s_t})) - {Q_\theta }({s_t},{a_t})]]
            \label{Eq:JPi}
        \end{align}

        To overcome the sensitivity of this hyperparameter $\alpha$, the same author proposed a method that can adjust the temperature parameter automatically \cite{haarnoja2018soft_2}.
        In their work, by using a dual objective approach, the problem has been formulated as a maximum entropy RL optimization problem with a minimum entropy constraint.
        In practice, the temperature parameter $\alpha$ is approximated by the neural network, given in Eq.~(\ref{Eq:Ja}):
        \begin{align}
        J_{\alpha}= {\mathbb{E}_{{a_t} \sim {\pi _t}}}[ - \alpha \log {\pi _t}({a_t}|{s_t}) - \alpha \bar H]
            \label{Eq:Ja}
        \end{align}
        where the $\bar{H}$ represents a desired minimum expected entropy.

\section{Design of One-To-One Air Combat Environment}\label{SE:Env}

In order to solve the optimization problem mentioned in  Eq.~(\ref{Eq:problemOfAirCombat}) by RL-based methods, we need an environment to interact with the agent.
In this section, we will introduce the setting of the environment includes the state space, the action space, the transition function, and the reward regulation function.
Also, we will introduce the way we translate the global state space into the relative state space, which can simplify the state information of the UCAVs a lot.

        \subsection{Action Space}
        The control law of the optimization above, called $u^b$ and $u^r$, can demonstrate as Eq.~(\ref{Eq:controlVec}).
        \begin{align}
            \begin{split}
                {u^b} &= [{\alpha ^b},{\mu^b},{\eta^b}]^ \mathrm{ T }\\
                {u^r} &= [{\alpha ^r},{\mu ^r},{\eta ^r}]^ \mathrm{ T }
            \end{split}
            \label{Eq:controlVec}
        \end{align}
        And the constraints of these control variables are given in Eq.~(\ref{Eq:controlConstraints}). So the action vectors can be designed as Eq.~(\ref{Eq:action}):
        \begin{align}
        \begin{split}
             a^b &= [\dot{\alpha^b},\dot{\mu^b}]\\
             a^r &= [\dot{\alpha^r},\dot{\mu^r}]
        \end{split}
        \label{Eq:action}
        \end{align} 
        
        By analysis, these control variables shown in Eq.~(\ref{Eq:controlVec}), at time step $t$, are  determined by the action vectors mentioned in Eq.~(\ref{Eq:action}) and the constraints shown in Eq.~(\ref{Eq:actionConstraint}).

        \begin{align}
            \begin{dcases}                                                                                          
                \dot \alpha (t) \in {\rm{[ - }}\Delta \alpha ,\Delta \alpha ]{\rm{       }}\\
                \alpha (t) = \int\limits_{{t_0}}^t {\dot \alpha (u)du}  \in {\rm{[}}{\alpha _{\min }},{\alpha _{\max }}{\rm{]  }}\\
                \dot \mu (t) \in [{\rm{ - }}\Delta \mu ,\Delta \mu ]\\
                \mu (t) = \int\limits_{{t_0}}^t {\dot \mu } (u)du \in [ - \pi ,\pi ]\\
            \end{dcases}
            \label{Eq:actionConstraint}
        \end{align}

        \subsection{State Space}
        A suitable design of the state space can reduce the burden of algorithms as well as accelerate the convergence speed. 
        The form of state can be designed as Eq.~(\ref{Eq:otherPaperS})\cite{kong2020uav}:
        \begin{align}
             {s} &= {[{x^b},{y^b},{z^b},{v^b},{\gamma ^b},{\chi ^b},{x^r},{y^r},{z^r},{v^r},{\gamma ^r},{\chi ^r}]^ \mathrm{ T }}
            \label{Eq:otherPaperS}
        \end{align}

\begin{figure}[ht]
            	\centering
            	\includegraphics[width = 2.5in]{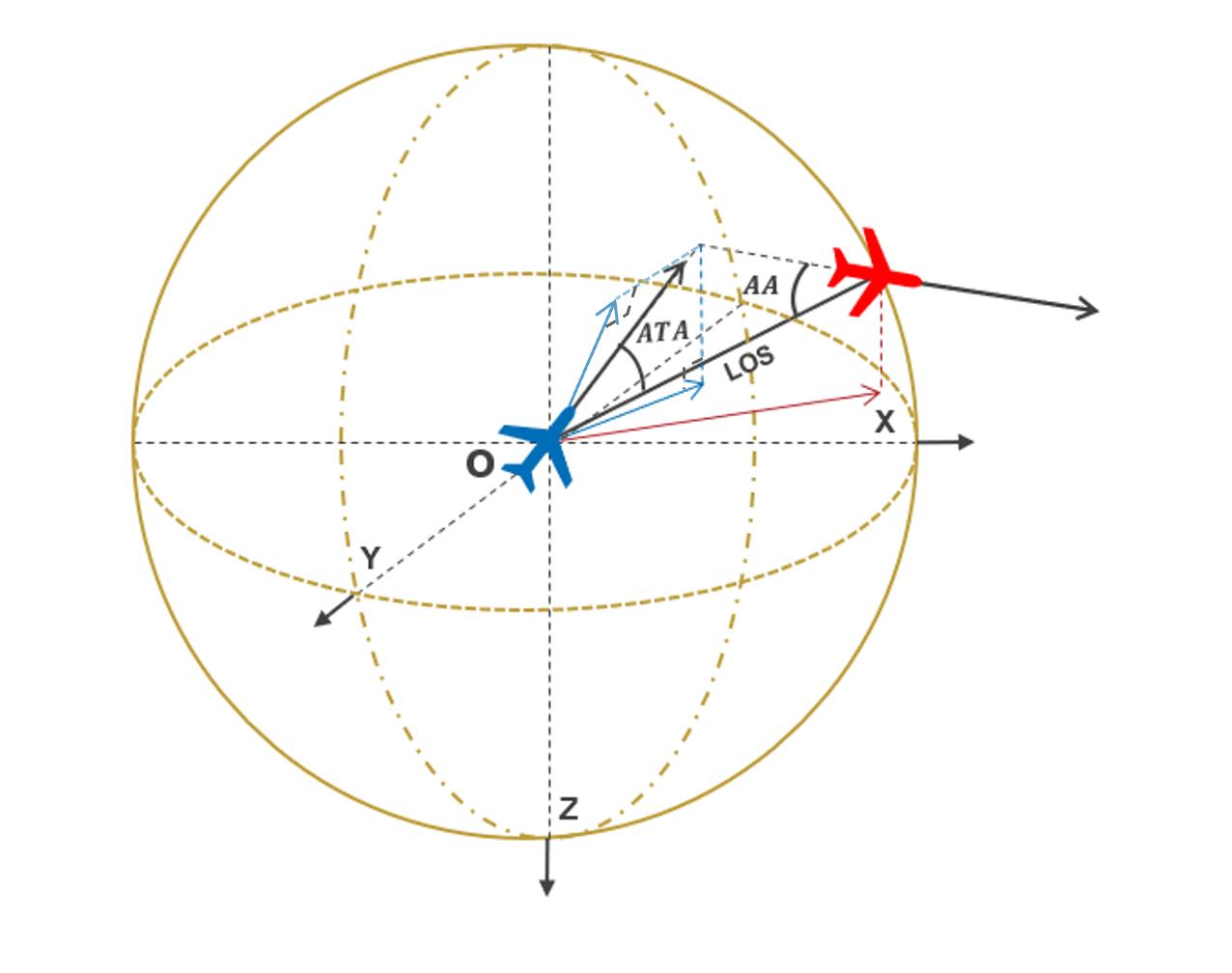}
            	\caption{The diagrammatic sketch of three dimensional air combat }
            	\label{Fig:3dcombat}
            \end{figure}

        To reduce the dimension of the state and preprocess the information of state, we translate the global state, which is mentioned in Eq.~(\ref{Eq:otherPaperS}), into the relative state, as shown in Eq.~({\ref{Eq:Stateb}}) and Eq.~({\ref{Eq:Stater}}).

        \begin{align}
        \begin{split}
               {s^b}&=[ATA^b_{XOY},ATA^b_{YOZ}, AA^b_{XOY},AA^b_{YOZ},
               \\& {\quad\mu ^b},{D_{LOS}},{v^b},{\gamma ^b},{v^r},{\alpha ^b},{z^b}]^ \mathrm{ T }
         \end{split}
        \label{Eq:Stateb}
        \end{align}
        
        \begin{align}
            \begin{split}
                s^r &=[{ATA^r_{XOY}},{ATA^r_{YOZ}}, {AA^r_{XOY}},{AA^r_{YOZ}},\\& \quad{\mu ^r},{D_{LOS}},{v^r},{\gamma ^r},{v^b},{\alpha ^r},{z^r}]^ \mathrm{ T }
            \end{split}
            \label{Eq:Stater}
        \end{align}
        Where the subscripts $XOY$ and $YOZ$ denote the projection of angle in $XOY$ and $YOZ$, as shown in Figure~\ref{Fig:3dcombat}.
         In addition, $D_{LOS}$ means the distance of these two UCAVs in air combat scenarios.
        We introduce relative variables to encode the naive state space in Eq.~(\ref{Eq:otherPaperS}), decoupling the relationship between the absolute position of UCAVs and the policy produced by the network, as given in Eq.~(\ref{Eq:Stateb}) and Eq.~(\ref{Eq:Stater}).
        
        We normalize the parameters in Eq.~(\ref{Eq:Stateb}) and Eq.~(\ref{Eq:Stater}) to improve the convergence of the neural network. 
        And the normalized state space vector are given in Eq.~(\ref{Eq:Stateb_n}) and Eq.~(\ref{Eq:Stater_n}), respectively.
         \begin{align}
        \begin{split}
               {s^b}&=\big[
               \frac{ATA^b_{XOY}}{2\pi},
               \frac{ATA^b_{YOZ}}{2\pi}, 
                \frac{AA^b_{XOY}}{2\pi},
              \frac{AA^b_{YOZ}}{2\pi},
                \frac{{\mu}^b}{2\pi},
                \frac{D_{LOS}}{D_{\max}},\\
               & \frac{v^b}{v_{\max}-v_{\min}},
                \frac{{\gamma}^b}{2\pi},
                \frac{v^r}{v_{\max}-v_{\min}} ,
               \frac{\alpha^b}{{\alpha}_{\max}-{\alpha}_{\min}},
                \frac{-z^b}{h_{\max}-h_{\min}}
                \big]^ \mathrm{ T }
         \end{split}
        \label{Eq:Stateb_n}
        \end{align}
         \begin{align}
        \begin{split}
               {s^r}&=\big[
               \frac{ATA^r_{XOY}}{2\pi},
               \frac{ATA^r_{YOZ}}{2\pi}, 
                \frac{AA^r_{XOY}}{2\pi},
              \frac{AA^r_{YOZ}}{2\pi},
                \frac{{\mu}^r}{2\pi},
                \frac{D_{LOS}}{D_{\max}},\\
               & \frac{v^r}{v_{\max}-v_{\min}},
                \frac{{\gamma}^r}{2\pi},
                \frac{v^r}{v_{\max}-v_{\min}} ,
               \frac{\alpha^r}{{\alpha}_{\max}-{\alpha}_{\min}},
                \frac{-z^r}{h_{\max}-h_{\min}}
                \big]^ \mathrm{ T }
         \end{split}
        \label{Eq:Stater_n}
        \end{align}

            

        \subsection{Transition Function}
        \begin{figure*}[h]
            	\centering
            	\includegraphics[width = 5.5in]{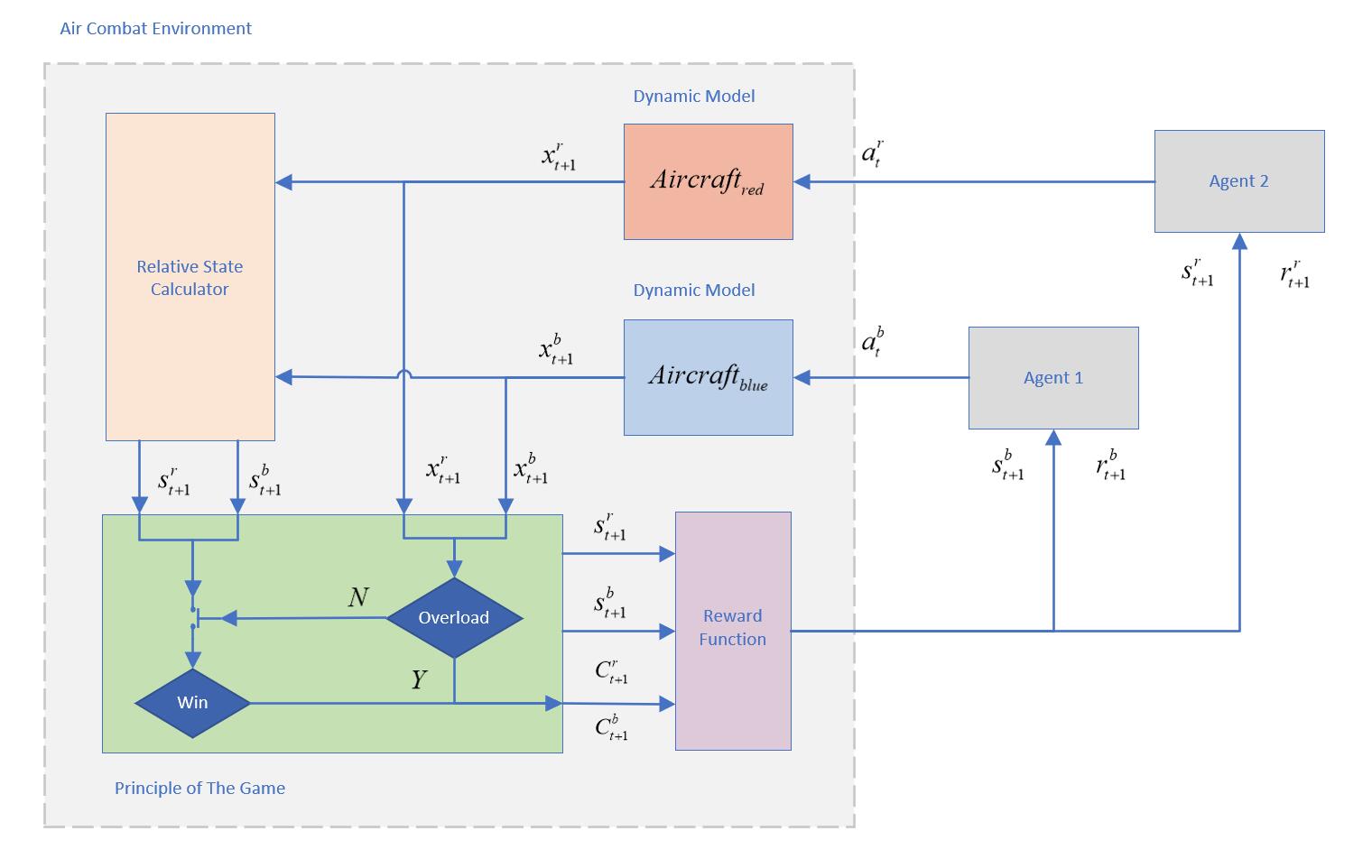}
            	\caption{The framework of the air combat environment }
            	\label{Fig:envFramework}
            \end{figure*}
        
         We use the first order approximation and  discretize the math model in section~\ref{SE:UCAV_dynamic_model}, as shown in Eq.~(\ref{Eq:firstOrder}).
        \begin{align}
           x(t + 1) = x(t) + \dot x(a(t))\Delta t 
           \label{Eq:firstOrder}
        \end{align}

        And the criterion we evaluating the situation of $\text{UCAV}^b$, denoted as $C^b$, in air combat scenario is given in Criterion~\ref{algo:airCombatLogic}.
            \renewcommand{\algorithmicrequire}{\textbf{Input:}}  
                \renewcommand{\algorithmicensure}{\textbf{Output:}} 
                \renewcommand{\thealgorithm}{1}
                \floatname{algorithm}{Criterion}
                \begin{algorithm}[hbt!]
                \setcounter{algorithm}{1}
                \caption{Air combat logic in the view of $\text{UCAV}^b$} 
                \begin{algorithmic}[1]
                \Require
                State:\;$s^b$
                \Ensure
                Situation of UCAV in air combat:\;$C^b$
                \item $D_{LOS}$: Euclidean distance between UCAV 
                \If{\quad$h^b\notin[h_{\min},h_{\max}]\; \text{or} \;n(\alpha,v)>10g \;\text{or}\; v^b\notin[v_{\min},v_{max}]$}
                    \State $C^b \quad=\quad overloaded $
                \ElsIf{\quad$|AA^r|<60^{\circ}$\;and\;$|ATA^r|<60^{\circ}$\;and\;$D_{LOS}\in [D_{\min},D_{\max}]$}
                    \State $ C^b \quad=\quad killed $
                \ElsIf{\quad$|AA^b|<60^{\circ}$\;and\;$|ATA^b|<60^{\circ}$\;and\;$D_{LOS}\in [D_{\min},D_{\max}]$}
                    \State $ C^b \quad=\quad win $
                \Else 
                    \State $ C^b \quad=\quad survival$
                \EndIf 
                \end{algorithmic}
                \label{algo:airCombatLogic}
                \end{algorithm}
              
            

        \subsection{Reward}
        The design of the reward function is the most significant step in RL.
        The basic task of the UCAV in air combat scenarios is to achieve the advantageous position as soon as possible, the reward function of $\text{UCAV}^b$ can be described as Eq.~(\ref{Eq:reward}):
        \begin{small}
        \begin{align}
        R = 
        \begin{cases}
        r_{1}\quad &\textbf{if}\;C^b\; ==\;win\\
        r_{2}    & \textbf{if}\;C^b  \; == \; survival\;\textbf{and}\;C^r  \; == \;overloaded\\
        r_{3}   &\textbf{if}\;C^b \; == \; overloaded\;\textbf{or}\;\;C^b \; ==\; killed  \\
        r_{4}     &\textbf{if}\;C^b   \; ==\; survival\;\textbf{and}\;C^r  \; ==\; survival\\
        \end{cases}
        \label{Eq:reward}
        \end{align}
        \end{small}
        In the air combat scenario, the UCAVs' situation is determined by Criterion~\ref{algo:airCombatLogic}.
        When $\text{UCAV}^b$ goes into the advantageous position, it will be given an absolute win reward $r_{1}$.
        If $\text{UCAV}^r$ overloads meanwhile $\text{UCAV}^b$ is still survival, $\text{UCAV}^b$ is also considered to win the game and awarded a relative win reward $ r_{2}$. 
        Punishment $ r_{3}$ will be given while the UCAV is killed or overloaded.
        To minimize the cumulative time, each UCAV is given the step punishment $r_{4}$ until the end of the game.

        The framework of the environment can be illustrated in Figure~\ref{Fig:envFramework}.
    
\section{Homotopy-based Soft Actor-Critic Method}\label{SE:method}

The agent can hardly learn a policy to achieve an advantageous position only via the sparse reward signal and random exploration. Meanwhile, using artificial prior knowledge may also bias the direction from the optimal policy.

To find a way out of this dilemma, we combine the advantages of the artificial prior knowledge and the sparse reward task formulation and propose a novel homotopy-based soft actor-critic algorithm.
Specifically, in the first stage, this algorithm uses the artificial prior knowledge to find a suboptimal policy. And then, by following a homotopy path in solution space, this algorithm guides the suboptimal policy to the optimal policy, gradually.

In this section, we first improve the RL method by the idea of homotopy. Furthermore, we prove the convergence and the feasibility of this method. Finally, we apply this method to the air combat scenario.

\subsection{Homotopy Based Reinforcement Learning Problem}\label{SE:HPFRL}

 The RL problem with sparse reward setting $R$ can be formalized as a nonlinear programming (NLP) problem\cite{bertsekas2019reinforcement}, as shown in \textbf{NLP1}. 
 
 \noindent\textbf{NLP1 (Original Problem)}
 \begin{align}
   \centering
    \begin{split}
        \mathop {\max }\limits_{\phi} F_1(\phi) =& 
        \mathbb{E}_{s_0\sim {\rho _0}({s_0})}
        \left[ {\sum\limits_{t = 0}^T {{\gamma^t}\big[{R}({s_t},{a_t})} \big]} \right]\\
        h_j(\phi)=&0 \qquad j=1,\cdot\cdot\cdot,m\\
        g_j(\phi)\ge&0 \qquad j=1,\cdot\cdot\cdot,s\\
    \end{split}
    \nonumber
    \end{align}
Where $a_t\sim\pi(\cdot|s_t)$ and $\phi$ means the parameter of actor network. 
$h(\cdot)=0$ and $g(\cdot)\ge0$ represent the equality constraints and inequality constraints of the environment, respectively. 
 
To deal with this sparse reward RL problem, artificial priors can be introduced as an extra reward $R^{extra}$, which can enhance the feedback on the quality of the agent's action each step, to guide agents to find a feasible policy.
With this extra reward $R^{extra}$, we calculate the total reward as $R+R^{extra}$.
And then, the RL problem with $R+R^{extra}$ can also be formalized as the NLP problem, as shown in \textbf{NLP2}.

\noindent\textbf{NLP2 (Auxiliary Problem)}
  \begin{align}
   \centering
    \begin{split}
        \mathop {\max }\limits_{\phi} F_2(\phi) = & 
        \mathbb{E}_{s_0\sim {\rho _0}({s_0})}
        \left[{\sum\limits_{t = 0}^T {{\gamma^t} \big[  {R}({s_t},{a_t})} +{R^{extra}}({s_t},{a_t}) \big]} \right]\\
        h_j(\phi)=&0 \qquad j=1,\cdot\cdot\cdot,m\\
        g_j(\phi)\ge&0 \qquad j=1,\cdot\cdot\cdot,s
    \end{split}
    \nonumber
    \end{align}
  
  However, this feasible policy may be a suboptimal policy and the artificial priors may distort the original problem.
 To make the policy converge to the optimal solution in sparse reward RL problem while keeping the optimization objective of the original problem. We propose a function with the idea of homotopy, given in Eq.~(\ref{Eq:_F3}).
 \begin{align}
 \begin{split}
     F_3(\phi,q)=(1-q)F_2(\phi)+qF_1(\phi)
 \end{split}
 \label{Eq:_F3}
 \end{align}
Where $q\in [0,1]$ is the weight of the auxiliary operator $F_2(\cdot)$ and the nonlinear operator $F_1(\cdot)$.
 Because the original problem (\textbf{NLP1}) and the auxiliary problem (\textbf{NLP2}) are only different in the design of the reward.
 With Eq.~(\ref{Eq:_F3}), a homotopy NLP problem can be formalized as shown in \textbf{NLP3}.
 
\noindent\textbf{NLP3 (Homotopy Problem)}
   \begin{align}
   \centering
    \begin{split}
        \mathop {\max }\limits_{\phi} F_3(\phi,q) = & (1-q)F_2(\phi)+qF_1(\phi)
        \\
        h_j(\phi)=&0 \qquad j=1,\cdot\cdot\cdot,m\\
        g_j(\phi)\ge&0 \qquad j=1,\cdot\cdot\cdot,s
    \end{split}
    \nonumber
    \end{align}
This modality has the character that when the variable $q$ gets 0 and 1 we could obtain the equations in Eq.~(\ref{Eq:q=0_1homotopyFun}), respectively:
\begin{align}
\begin{split}
F_3[\phi,q]\big|{}_{q = 0} &= F_2[\phi]  \\
F_3[\phi,q]\big|{}_{q = 1} &= F_1[\phi]  
\end{split}
\label{Eq:q=0_1homotopyFun}
\end{align}

So if the variable $q$ is 0, the homotopy problem (\textbf{NLP3}) is equivalent to the auxiliary problem (\textbf{NLP2}). Also when the variable $q$ gets 1, the homotopy problem (\textbf{NLP3}) is equivalent to the original problem (\textbf{NLP1}).
Ideally, with the weight $q$ changes from 0 to 1, the solution of \textbf{NLP3} can also change from the solution of \textbf{NLP2} to the solution of \textbf{NLP1} which we would like to solve, continuously. This continuous and variable process is called homotopy.

We design a homotopy reward function as shown in Eq.~(\ref{Eq:homotopyReward}).
 \begin{align}
        R^{\text{H}}(q)=q(R+R^{extra})+(1-q)R
        \label{Eq:homotopyReward}
    \end{align}

With Theorem~\ref{Th:equal}, we can find that $\mathbf{NLP3}$ is equivalent to the RL problem which aims to maximize the exception of accumulated $R^{\text{H}}$.
\begin{theorem}
$F_3(\cdot)$ equals to the exception of accumulated homotopy reward as shown in Eq.~(\ref{Eq:F3_rl}).
\begin{align}
   \centering
    \begin{split}
        F_3(\phi,q) = & 
        \mathbb{E}_{s_0\sim {\rho _0}({s_0})}
        \left[{\sum\limits_{t = 0}^T {{\gamma^t} \big[ } R^{\text{H}}({s_t},{a_t},q) \big]} \right]\\
    \end{split}
    \label{Eq:F3_rl}
    \end{align}
    \label{Th:equal}
\end{theorem}   
Theorem~\ref{Th:equal} is proved in ~\ref{Appendix:A}.

\subsection{Homotopy Path Following with Predictor-Corrector Method}\label{SE:HPFPCM}
In Section~\ref{SE:HPFRL}, we design a function with the idea of homotopy to solve the convergence difficulty of the original problem (\textbf{NLP1}).

 Theorem~\ref{Th:PE} ensures that the solution of the auxiliary problem (\textbf{NLP2}) can transform to the solution of the original problem (\textbf{NLP1}) via $F_3(\cdot)$.

\begin{theorem}
The homotopy path between the solution of \textbf{NLP2} and the solution of \textbf{NLP1} must exist.
\label{Th:PE}
\end{theorem}
 Where the "path" means a piecewise differentiable curve in solution space, and the proof of homotopy path existence is given in ~\ref{Appendix:B}.


Then, in this section, the idea of the traditional method called horizontal corrector and elevator predictor\cite{lemke1984pathways} is used to guide a feasible solution to the optimal solution of the original problem ($\mathbf{NLP1}$), following the homotopy path in solution space.

Firstly, we denote the task of an auxiliary problem ($\textbf{NLP2}$) as $\mathcal{M}_0$ and use $\mathcal{M}_T$ to represent the task of the original problem ($\textbf{NLP1}$). 
The goal of $\mathcal{M}_0$ is to find the optimal policy, which is denoted as $\pi^*_{\mathcal{M}_0}(\phi)$, to maximize the accumulative reward, namely the solution of the auxiliary problem (\textbf{NLP2}).
Theoretically,
if we can follow a feasible homotopy path,
the optimal policy of the task $\mathcal{M}_T$, which is represented as $\pi^*_{\mathcal{M}_T}(\phi)$,
can also be found.

Then, we denote the number of the steps that $q$ needs to complete the transition of the solution as $N$, namely the change of $q$ can be represented as a series of values $q_0,q_1,q_2,q_3,\cdot\cdot\cdot,q_N$ satisfying $q_0=0<q_1<q_2<q_3<\cdot\cdot\cdot<q_N=1$.
This allow us to define a sequence of corresponding sub-tasks as $\mathcal{M}_0,\mathcal{M}_1,\cdot\cdot\cdot,\mathcal{M}_N$, where $\mathcal{M}_N=\mathcal{M}_T$.
Also, we denote the tuple of the parameter of actor $\phi$ and the value of $q$ in step $n\in\{0,1,2,3\cdot\cdot\cdot N\}$ as $y_n=(\phi_n,q_n)$.

\noindent $\textbf{Predictor}$
    
    In the predictor step $n$, 
    if the solution $\pi^*_{\mathcal{M}_n}(\phi)$ of the task $\mathcal{M}_n$ has been calculated, we give a predictor of the parameter $\phi^*_{n+1}$ of the next optimal policy$\pi^*_{\mathcal{M}_{n+1}}(\phi)$ along the homotopy path.
    
    Here, for the convenience calculation. we use the idea of elevator predictor\cite{lemke1984pathways} to predict the solution of next task  $\mathcal{M}_{n+1}$,
    as shown in Eq.~(\ref{Eq:elevatorPredictor})
    \begin{align}
        \hat{y}_{n+1}=(\phi^*_{n},q_{n+1})
        \label{Eq:elevatorPredictor}
    \end{align}

\noindent $\textbf{Corrector}$

    After the predictor step to the new task $\mathcal{M}_{n+1}$, the predicted parameter tuple $\hat{y}_{n+1}$ may drift away from the homotopy path.
    The corrector step aims to get the solution back onto or very close to the path. Namely correct the predicted tuple $\hat{y}_{n+1}$ to the real solution tuple $y_{n+1}$.
    This corrector is called a horizontal corrector because the parameter $q_{n+1}$ will not be changed during correcting.
    
    In this process, We only need to verify if the solution satisfies the accuracy demands. Because the RL-based methods naturally have the ability of corrector via gradient descent.
    The convergence criterion can be described as shown in Eq.~(\ref{Eq:adjustmentOfH}).
    On account of the homotopy problem ($\mathbf{NLP3}$) has been transformed into the formalization $H(y_n)=0$.
    And the relative proof is given in ~\ref{Appendix:B}.
    \begin{align}
        || H(y_n) ||\le \varepsilon
        \label{Eq:adjustmentOfH}
    \end{align}
    Where $\varepsilon$ is a threshold value of this convergence criterion. 

\subsection{Algorithm}\label{SE:HSAC}

We propose the Theorem~\ref{convergence}:
\begin{theorem}
When the weight $q_n$ transform from 0 to 1, via horizontal corrector elevator predictor method, the solution $\pi^*_{\mathcal{M}_n}(\phi)$ of task $\mathcal{M}_n$ could converge to the solution $\pi^*_{\mathcal{M}_T}(\phi)$ of task $\mathcal{M}_T$.
\label{convergence}
\end{theorem}
Where the proof of the convergence of the predictor-corrector path-following method is given in \ref{Appendix:B}.

On account of this theorem, a homotopy-based soft actor-critic algorithm can be designed as Algorithm ~\ref{algo:HSAC}.

\renewcommand{\algorithmicrequire}{\textbf{Input:}}
\renewcommand{\algorithmicensure}{\textbf{Output:}} 
\renewcommand{\thealgorithm}{1}
\floatname{algorithm}{Algorithm}
\begin{algorithm*}[hbt]
\caption{Homotopy-based Soft Actor-Critic} 
\begin{algorithmic}[1]
\Require
Initialized Parameters:\{$\theta_1$,\;$\theta_2$,\;$\phi$,\;$M$,\;$N$\}
\Ensure
Optimized parameters: \{$\theta_1$,\;$\theta_2$,\;$\phi$\}
\item \textbf{Initialize target network weights:}
\item $\bar{\theta_1}\gets\theta_1$,\quad$\bar{\theta_2}\gets\theta_2$,\quad$\varepsilon>0$
\item \textbf{Initialize the auxiliary weight:}
\item $q_0\gets1,\; n\gets0$
\item\textbf{Empty the Replay Buffer:}
\item $\mathcal{D}\gets\emptyset$
\item\textbf{Empty the gradient of policy $\nabla\pi$ Buffer:}
\item $\mathcal{P}\gets\emptyset,\mathcal{X}\gets{[1,2,3,\cdot\cdot\cdot,M]^{\mathrm{T}}}$
\item $\mathcal{X}\gets{Add A Column With Ones (\mathcal{X})}$
\For {each episode}
\For{each step in the episode}
\State $a_t\sim\pi_{\phi}(\cdot \big| s_t) $\qquad\
\State $s_{t+1},R^{\text{H}}_{t+1}(k_n) \sim T(\cdot\big|s_t,a_t)$\;
\State $\mathcal{D}\gets\mathcal{D}\cup(s_t,a_t,r_{t+1},s_{t+1})$  
\EndFor
\For{each update step}
\State\emph{\textbf{Horizontal Corrector:}}
\State$\theta_i\gets\theta_i-\lambda_Q\hat{\nabla}_{\theta_i}J_Q(\theta_i)$\;for\;$i\in\{1,2\}$ 
\State$\phi\gets\phi-\lambda_{\pi}\hat{\nabla}_{\phi}J_{\pi}(\phi,\mathop {\arg \min }\limits_{\theta_i}  Q_{\theta_i})$
\State$\alpha\gets\alpha-\lambda_{\alpha}\hat{\nabla}_{\alpha}J_{\alpha}(\alpha)$
\State $\bar{\theta_i}\gets\tau\theta_i+(1-\tau)\bar{\theta_i}\;for\;i\in\{1,2\}$
\State
$\mathcal{P}\gets\mathcal{P}\cup\hat{\nabla}_{\phi}J_{\pi}$
\State\emph{\textbf{Elevator Predictor:}}
\If{ $len(\mathcal{P})>=M$}
\State$k_{{\nabla\pi}}\gets{(\mathcal{X}^\mathrm{T}\mathcal{X})^{-1}\mathcal{X}^\mathrm{T}\mathcal{P}}$
\If{$\big|k_{{\nabla\pi}}\big|<\varepsilon\;\textbf{and}\;q_n>0$}
\State $q_{n+1}\gets{q_{n}-\frac{1}{N}}$
\State $n\gets n+1$
\State $\mathcal{P}\gets\emptyset$
\EndIf
\EndIf
\EndFor
\EndFor
\end{algorithmic}
\label{algo:HSAC}
\end{algorithm*}

     \begin{figure*}[hbt]
        \centering
        \includegraphics[width = 6in]{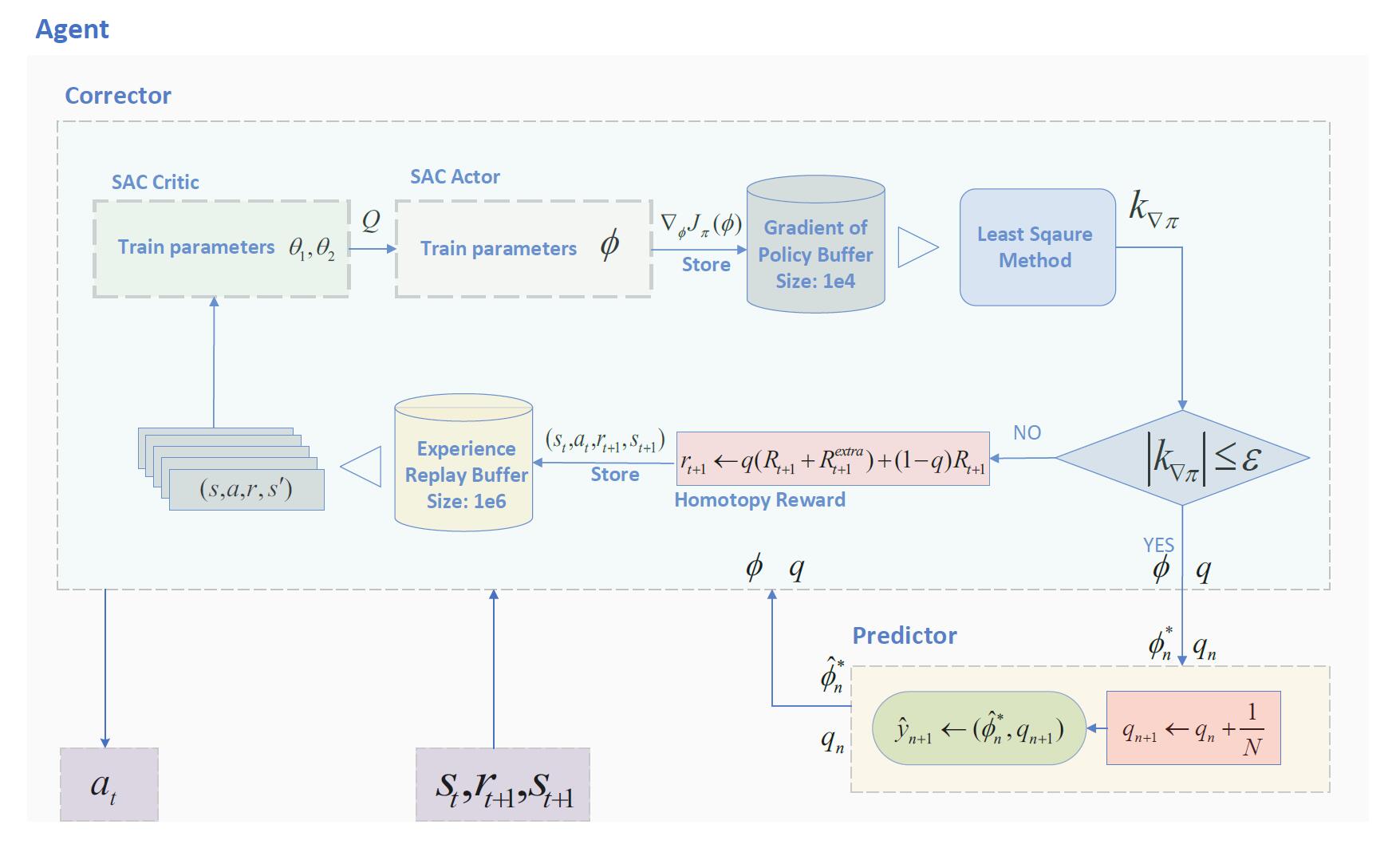}
        \caption{The construction of the HSAC method}
        \label{Fig:agent}
        \end{figure*}
Practically, we use the variation tendency of parameter $\phi$ in the 
actor network of SAC algorithm, denoted as $k_{\nabla \pi}$, to verify if the policy is converged to the optimal policy of $\mathcal{M}_n$.
To calculate the slope $k_{\nabla \pi}$ of the policy gradient changes, we fit the data of the policy gradient as a first-order function by the least square method.
And the slope $k_{\nabla \pi}$ is used to estimate the quality of convergence in task $\mathcal{M}_n$.
More details are given in Algorithm~\ref{algo:HSAC}.

In Algorithm~\ref{algo:HSAC},  $\mathcal{P}$ is the buffer to store the gradient of policy $\nabla\pi$ for calculating the slope $k_{\nabla \pi}$. 
$M$ is the sample size, a parameter of the least square method. 
$N$ means the total number of iteration that the homotopy method needs.

In the horizontal corrector step, the parameter $\phi$ of the policy $\pi_{\mathcal{M}_n}(\phi)$ is iterated to correct the policy to approach the optimal policy $\pi^*_{\mathcal{M}_n}(\phi)$ during the task $\mathcal{M}_n$.
If the slope $k_{\nabla \pi}$ satisfies the demand $|k_{\nabla \pi}|<\varepsilon$, the $n^{th}$ corrector step is finished. Otherwise iterate the parameter $\phi$ continuously.

In the elevator predictor step, firstly clear the policy gradient buffer $\mathcal{P}$ to store the new data in the new task $\mathcal{M}_{n+1}$. 
Then clone the parameter $\phi$ of the optimal policy $\pi^*_{\mathcal{M}_n}(\phi)$ in the last task $\mathcal{M}_{n}$ directly to the policy of new task $\mathcal{M}_{n+1}$.

Through this method, at the beginning of the training process, the policy can easily converge to a feasible policy for the task $\mathcal{M}_{0}$ with the help of extra reward $R^{extra}$.
Then the negative influences from extra reward $R^{extra}$ will be ablated gradually, with the iterations via the corrector-predictor method.
Finally, the optimal policy of the target task $\mathcal{M}_{T}$ could be acquired, along with the weight $q$ transit from 0 to 1. 

\subsection{The Application of HSAC in Air Combat}\label{SE:apply}

In the air combat scenario, we need to find an equilibrium point of the two-target differential game, mentioned in Eq.~(\ref{Eq:problemOfAirCombat}), to ensure the quality of the policy.
Here, we use the idea of self-play to simplify the two-target differential game as a self-play RL problem.
Namely, both $\text{UCAV}^b$ and $\text{UCAV}^r$ use the same policy in the confrontation task. 
And this problem can be converted to the RL problem which only needs to learn a policy to maximize the accumulated reward just like the form of \textbf{NLP1}. 

According to the HSAC method, firstly, we define the simple reward $R$ as mentioned in Eq.~(\ref{Eq:reward})
and formalize the original problem as shown in \textbf{NLP1}.
The equality constraints mentioned in Eq.~(\ref{Eq:3-dmotionModel}), Eq.~(\ref{Eq:pointMassModel}), and Eq.~(\ref{Eq:liftAndDragForce}) can be simplified as $h(\cdot)=0$.
Meanwhile, the inequality constraints mentioned in Eq.~(\ref{Eq:controlConstraints}) and Eq.~(\ref{Eq:dPLimit}) can be simplified as $g(\cdot)\ge0$.

Secondly, design the extra reward $R^{extra}$ as given in Eq.~(\ref{Eq:rExtra}) to create the auxiliary operator $F_2(\cdot)$.
And then solve the original problem with the HSAC algorithm. 
        \begin{align}
               {R^{extra}} =  
               \begin{cases}
                - {\Phi ^\mathrm{ T }}Q\Phi \quad\quad\quad\quad\quad\quad\quad\quad\;\quad\quad\quad\quad{\rm{ if}}\;{D_{LOS}} > {D_{\max }}\\
                - {\Phi ^\mathrm{ T }}Q\Phi  - k\left( {{{(\frac{{2{D_{LOS}}}}{{{D_{\max }} + {D_{\min }}}} - 1)}^2} + 1} \right)\;{{\rm{elif}} \quad {D_{LOS}} \le {D_{\max }}}
                \end{cases} 
                \label{Eq:rExtra}
        \end{align}
Where $\Phi=[ATA^b_{XOY},ATA^b_{YOZ},AA^b_{XOY},AA^b_{YOZ}]^\mathrm{ T }$ is the vector of relative angle $ATA^b$ and $AA^b$, in time $t$.
$Q\in\mathbb{R}^{3 \times 3 }$ is a positive diagonal matrix of weights and $k$ is the penalty coefficient of relative distance. 
Only when the relative distance of two UCAVs is smaller than the maximum attack range $D_{\max}$, the influence of relative distance will be considered in $R^{extra}$.

\section{Simulation}\label{SE:Sim}
 In this section, we are going to demonstrate the superiority of our method in sparse reward RL problems such as air combat via two different experiments.
 In the first experiment, we designed a simple task, namely attack horizontal flight UCAV task, to prove the advantage of the HSAC method. 
 In the second experiment, we want to prove the advantage of the HSAC method combined with the idea of self-play via the complex confrontation task.
 
 For the sake of distinction, we denote the method which only uses $R$ in the SAC algorithm as SAC-s, the method using combined reward $R+R^{extra}$ in the SAC algorithm as SAC-r.

 \subsection{Simulation Platform}
 
\begin{figure}[H]
        \centering
        \includegraphics[width = 2.75in]{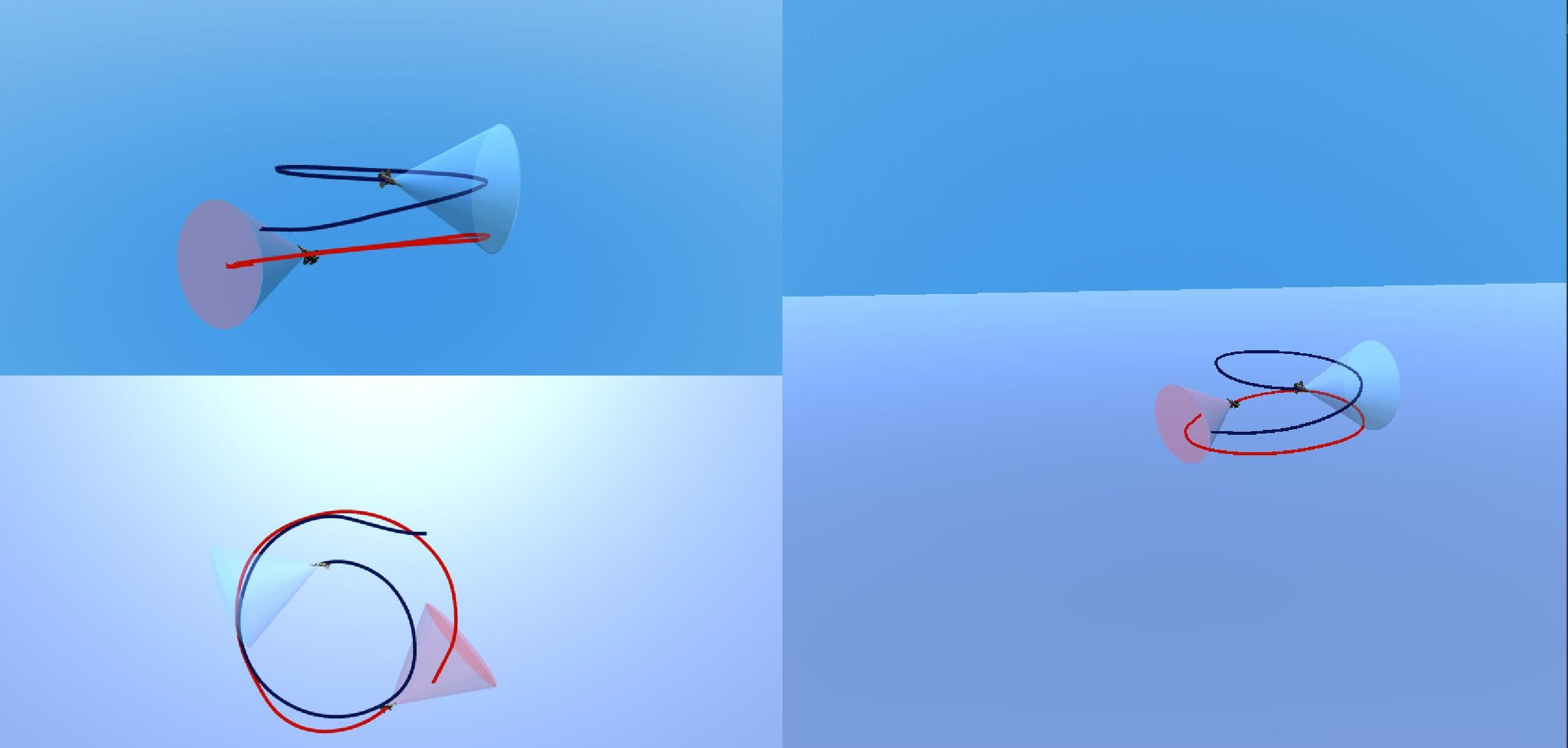}
        \caption{The simulation platform established by Unity3D and Gym }
        \label{Fig:unity}
        \end{figure}

  In this work, the environment is established by the framework of Gym\cite{1606.01540} using Python, rendered by Unity3D. And we use socket technology to realize the data interaction between Gym\cite{1606.01540} and Unity3D. 
 The interface of the platform is exhibited in Figure~\ref{Fig:unity}.

The agent is trained with an NVIDIA GeForce GTX 2080TI graphic card for PyTorch acceleration, and the framework of Algorithm~\ref{algo:HSAC} that the agent used is shown in Figure~\ref{Fig:agent}. 

The parameters of the air combat model, mentioned in Section~\ref{SE:UCAV_dynamic_model} and Section~\ref{SE:Air_combat_model}, are shown in Table~\ref{tb:aircombatpar}. And the hyperparameters of Algorithm~\ref{algo:HSAC} are shown in Table ~\ref{tb:HSACParameters}.

\subsection{Attack Horizontal Flight UCAV}\label{SE:AttackHorizontal}
\subsubsection{Task Setting}\label{SE:AHFATS}
This experiment is designed to compare the convergence of different methods in a simple task.

In the task of attack horizontal flight UCAV, the $\text{UCAV}^{r}$ performs horizontal flight maneuver with a random initial position and random initial heading angle. Meanwhile, $\text{UCAV}^b$, namely the agent we trained, aims to attack $\text{UCAV}^r$.

 \begin{figure}[hbt]
\centering
\includegraphics[width = 3in]{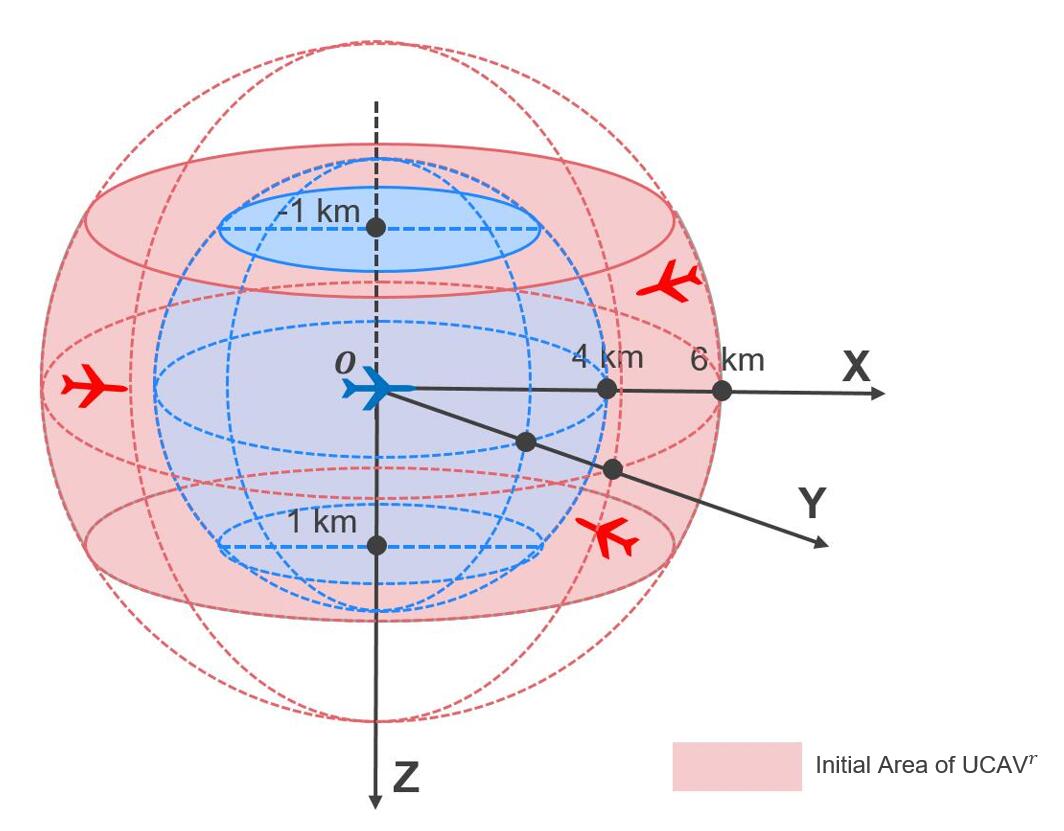}
\caption{The initial state of blue and red UCAV}
\label{Fig:InitialState}
\end{figure}
The initial position of $\text{UCAV}^b$ is the point $o$ in Figure~\ref{Fig:InitialState}, with 5 kilometers height.
And its initial heading angle is zero, namely the direction of $X$ axis.
To simulate radar detection of $\text{UCAV}^r$, the initial position of $\text{UCAV}^r$ has been set in the initial area as shown in Figure~\ref{Fig:InitialState}. 
The initial relative distance between $\text{UCAV}^r$ and $\text{UCAV}^b$ is in the range of 4km to 6km. The initial relative height is between -1km to 1km, and the initial heading angle of red UCAV is a random value from $-180^{\circ}$ to $180^{\circ}$. 

\subsubsection{Training Process}

To test the convergence of the methods and the performance in the original problem, We analyze from two perspectives, namely the training process and the evaluating process.
In the training process, the episode reward contains the extra reward, such as $R^{extra}$ and $R^{\text{H}}$, except the original reward $R$.
But in the evaluating process, the episode reward only includes the original reward $R$.
In this task, we evaluate the quality of the policy every ten training episodes.

\begin{figure}[hbt]
       \centering
        \begin{subfigure}[t]{8cm}
        \includegraphics[width = 3.8in]{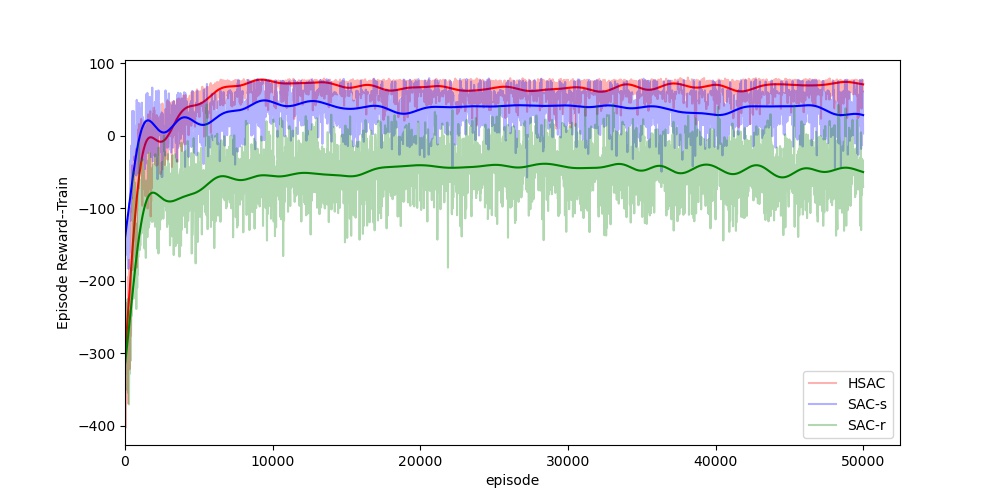}
        \caption{Episode reward in training process}
        \label{Fig:levelFlyTaska}
         \end{subfigure}
         \\
         \begin{subfigure}[t]{8cm}
        \includegraphics[width = 3.8in]{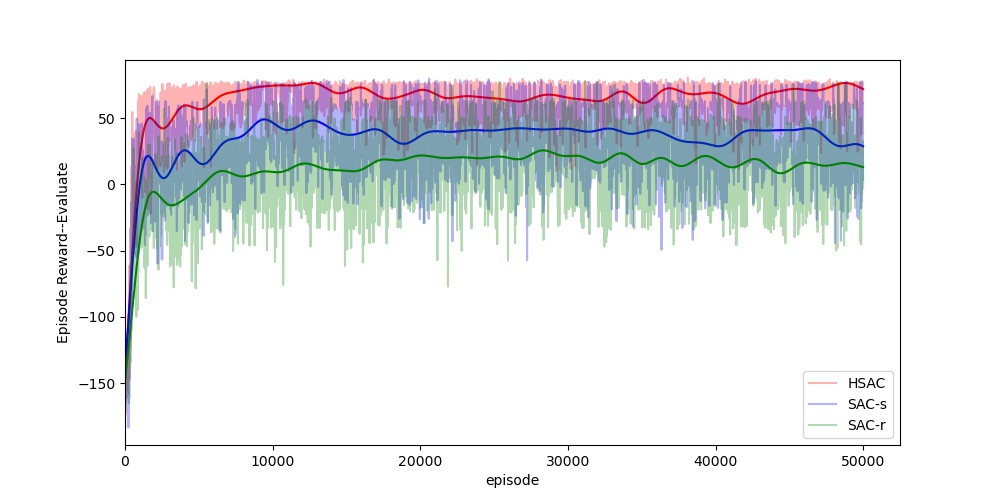}
        \caption{Episode reward in evaluating process}
        \label{Fig:levelFlyTaskb}
         \end{subfigure}\\
        
         \caption{The episode reward of different methods in attack horizontal flight UCAV task, the results of training process reflect the quality of convergence, and the results of evaluating process reflect the performance of different methods in the original problem. }
    \label{Fig:levelFlyTask}
 \end{figure}

We train the models in the first task with $5\times10^4$ episodes, as shown in Figure~\ref{Fig:levelFlyTask}.
The changes of episode reward in the training process via different methods are shown in Figure~\ref{Fig:levelFlyTaska}.
The changes of episode reward in the evaluating process via different methods are shown in Figure~\ref{Fig:levelFlyTaskb}.

In Figure~\ref{Fig:levelFlyTaska}, it can be seen that the episode rewards in the training process are rising along with the iteration, which means that the agents with different methods acquire the ability to complete tasks gradually.
The reason why the agent with SAC-r gets lower episode rewards in Figure~\ref{Fig:levelFlyTaska} is that $R^{extra}$ has been designed less than zero.

To evaluate the quality of each agent during training, we need to focus on Figure~\ref{Fig:levelFlyTaskb}.
In Figure~\ref{Fig:levelFlyTaskb}, it is obvious that the agent using HSAC can achieve the highest episode reward in this task, compare with the agents trained by SAC-s and SAC-r.

Although the agent trained by SAC-r gets a lower episode reward than the agent trained by SAC-s, our method with the same artificial priors can still perform better than the other two methods.
Through these, we can find HSAC is insensitive to the extra reward and it can solve the challenge from sparse reward RL problem with the help of $R^{\text{H}}$, meanwhile, can find the optimal policy of the original problem.

Then we will compare the performance of these methods in different air combat scenarios, after training.
\subsubsection{Simulation Results}
\begin{figure}[hbt]
    \centering
    \includegraphics[width = 3.5in]{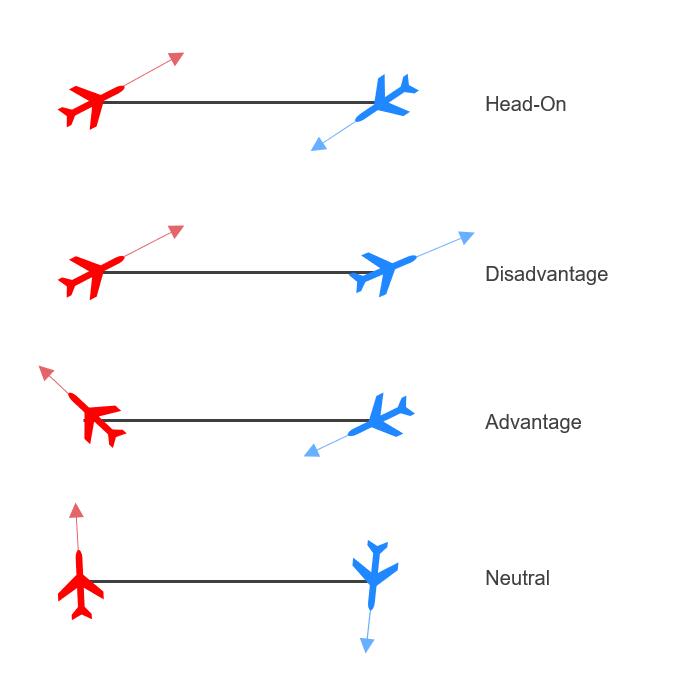}
    \caption{Four initial air combat scenario categories}
    \label{Fig:fourAircombaSituation}
\end{figure}

The initial state space of air combat geometry can be divided into four typical categories\cite{wang2020improving}(from the perspective of $\text{UCAV}^b$): head-on, neutral, disadvantageous, and advantageous, as shown in Figure~\ref{Fig:fourAircombaSituation}. 
In the different initial scenarios, the win probability of $\text{UCAV}^b$ will be disparate. 
For example, when $\text{UCAV}^b$ is in an advantageous position, it may be more likely to win the game, while $\text{UCAV}^b$ in the disadvantageous initial scenario is reversed. 
In neutral and head-one initial scenarios, the possibility of each side winning the game are similar.
Here, we will verify the performance of each well-trained agent in these four initial situations.

We selected four typical initial states in these four typical categories, more details are shown in Table~\ref{tb:4initialS}.
The initial bank angle $\mu$ and initial path angle $\gamma$ are set as zero for both $\text{UCAV}^b$ and $\text{UCAV}^r$. 

The performance of different agents with different methods are shown in Figure~\ref{Fig:DAIDS} and Table~\ref{tb:performsnceofAgents}. 
The data in Table~\ref{tb:performsnceofAgents} is produced by these agents with $10^4$ episodes evaluation.
The win rate of $\text{UCAV}^b$ and average time cost per episode in $10^4$ episodes are calculated to assess the qualities of each method.

\begin{table}[hbt!]
\centering
\scalebox{1.2}{
\begin{tabular}{| m{1.2cm}<{\centering}m{2cm}<{\centering} | m{1.8cm}<{\centering} | m{1.8cm}<{\centering} |m{1.8cm}<{\centering}|} 
            \hline
            \diagbox {Initial State}{Method}&& SAC-s & SAC-r & HSAC \\ \hline
            \multicolumn{1}{|l|}{Advantageous} & Win Rate &$\bm{100.00\% }$& $\bm{100.00 \%}$  & $\bm{ 100.00\%}$ \\ \cline{2-5} 
            \multicolumn{1}{|l|}{}    & Average Time Cost &  $ 42.07 s$   &  $\bm {37.90 s}$ & $39.40 s$\\ \hline
            
            \multicolumn{1}{|l|}{Disadvantageous} & Win Rate& $6.61\%$   &  $0.00\%$ &  $\bm{98.38\%}$ \\ \cline{2-5} 
            \multicolumn{1}{|l|}{}    & Average Time Cost &  $44.66 s$ &  $\bm{26.23 s}$  &  $35.91 s$ \\ \hline
            
            \multicolumn{1}{|l|}{Head-On} & Win Rate &   $73.32\%$& $7.43\% $ & $\bm{99.94\%}$  \\ \cline{2-5} 
            \multicolumn{1}{|l|}{}    & Average Time Cost & $39.15 s$ & $31.02 s$  & $\bm{24.73 s}$ \\ \hline
            
            \multicolumn{1}{|l|}{Neutral} & Win Rate &   $\bm{100.00\%}$& $3.66\%$  & $\bm{100.00\%} $ \\ \cline{2-5} 
            \multicolumn{1}{|l|}{}    & Average Time Cost &  $40.05 s$ &$22.57 s$  & $\bm{33.20 s}$ \\ \hline
            \end{tabular}

}
            \caption{Performance of different method trained agents in attack horizontal flight task with different initial states, after $10^4$ episodes evaluation.
             }
            \label{tb:performsnceofAgents}
\end{table}

As shown in Table~\ref{tb:performsnceofAgents}, agents trained by SAC-s, SAC-r and HSAC can all complete this task when the agent's initial state is in an advantageous situation.
In this situation, the agent using SAC-r can complete this task faster than other agents. 
However, in the other three initial situations, the performance of the agent trained by HSAC contrasts sharply with others.
In disadvantageous and head-on initial situations, the agent using HSAC has the highest probability to complete this task.
In a disadvantageous initial situation, the agent using SAC-s only has $6.61\%$ possibility to complete the task, meanwhile, the agent trained by SAC-r cannot complete the task.
In a head-on initial situation, the agent trained by SAC-s has a $73.32\%$ chance of completing this task, however, the agent trained by SAC-r only has a $7.43\%$ possibility to complete the task.
Although the agent trained by SAC-s as well as the agent trained by HSAC have the same win rate in a neutral initial situation,
the agent trained by HSAC can complete the task faster than the agent trained by SAC-s.
 However, the agent trained by SAC-r can hardly fulfill the task in a neutral initial situation.

In conclusion, SAC-s can describe the original task without bias, but we may face the challenge from sparse reward RL problem through this method.
SAC-r only performs well in the advantageous initial situation, in this experiment, because the introduced extra reward biases the converging direction of the original task.
However, the HSAC method we proposed, combining the advantages of both SAC-s and SAC-r, performs well in all initial situations.
Compare with SAC-s and SAC-r, HSAC has an overwhelming margin.

     \subsection{Confrontation Task}
     \subsubsection{Confrontation Task setting}

\begin{figure}[!hbt]
                \centering
                \includegraphics[width = 2.8in]{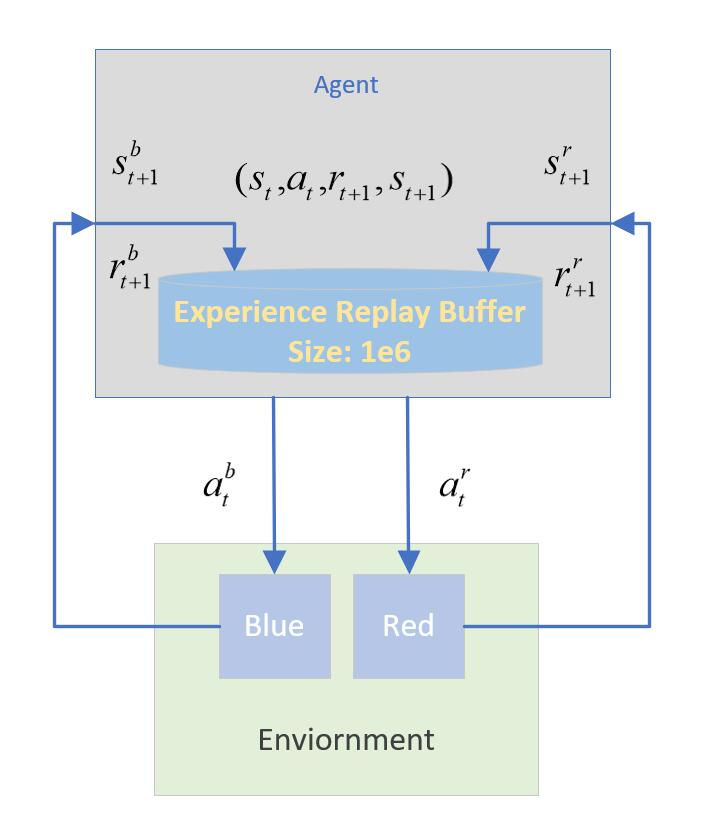}
                \caption{The structure of the confrontation task using the idea of self-play }
                \label{Fig:selfplayjpg}
                \end{figure}

       In this experiment, we would like to demonstrate the excellent performance of HSAC combined with the idea of self-play in the confrontation task.
       We combine the idea of self-play and HSAC, as mentioned in Section~\ref{SE:apply}, to train a more intelligent policy of UCAV in air combat, where optimize the policy via the pressure from the same policy.
       The structure of the self-play training process is shown in Figure~\ref{Fig:selfplayjpg},  where $\text{UCAV}^b$ and $\text{UCAV}^r$ share the same actor network and critic network.
        Meanwhile, the experience replay buffer and gradient of policy buffer needed in HSAC are shared to $\text{UCAV}^b$ and $\text{UCAV}^r$.
        And we follow the same initialization as the setting of attack horizontal flight UCAV task mentioned in Section~\ref{SE:AHFATS}.

\begin{table*}[hbt]
\centering
\scalebox{1.1}{
\begin{tabular}{|cc|cc|cc|cl|}
\hline
\multicolumn{2}{|c|}{\multirow{2}{*}{}}        & \multicolumn{2}{c|}{SAC-r} & \multicolumn{2}{c|}{SAC-s} & \multicolumn{2}{c|}{HSAC} \\ \cline{3-8} 
\multicolumn{2}{|c|}{}                         & \multicolumn{1}{c|}{Blue}  & Red  & \multicolumn{1}{c|}{Blue}     & Red     & \multicolumn{1}{c|}{Blue}   & Red   \\ \hline
\multicolumn{1}{|c|}{\multirow{2}{*}{\begin{tabular}[c]{@{}c@{}}Quality\\ of Policy\end{tabular}}} &
  Episode Reward &
  \multicolumn{1}{c|}{\textbf{-217.64}} &
  \textbf{111.72} &
  \multicolumn{1}{c|}{-1469.09} &
  -1339.13 &
  \multicolumn{1}{c|}{-387.52} &
  -382.57 \\ \cline{2-8} 
\multicolumn{1}{|c|}{}    & Difference Value    & \multicolumn{2}{c|}{329.36}       & \multicolumn{2}{c|}{129.96}             & \multicolumn{2}{c|}{\textbf{4.95}}           \\ \hline
\multicolumn{2}{|c|}{Convergence Episode Cost} & \multicolumn{2}{c|}{8250}         & \multicolumn{2}{c|}{6675}               & \multicolumn{2}{c|}{\textbf{4455}}           \\ \hline
\end{tabular}
}
\caption{The convergence episode cost for three different methods, the episode reward of the converged policy, and the difference value of converged episode reward between $\text{UCAV}^b$ and $\text{UCAV}^r$ in the self-play training process}
\label{Tb:SPConverge}
\end{table*}

        After training with self-play, we will compare the performance of the agents trained by SAC-s, SAC-r, and HSAC via confrontations experiments.
        
        \subsubsection{Training Process}

\begin{figure}[hbt]
\centering
\begin{subfigure}[t]{8cm}
\includegraphics[width = 3.8in]{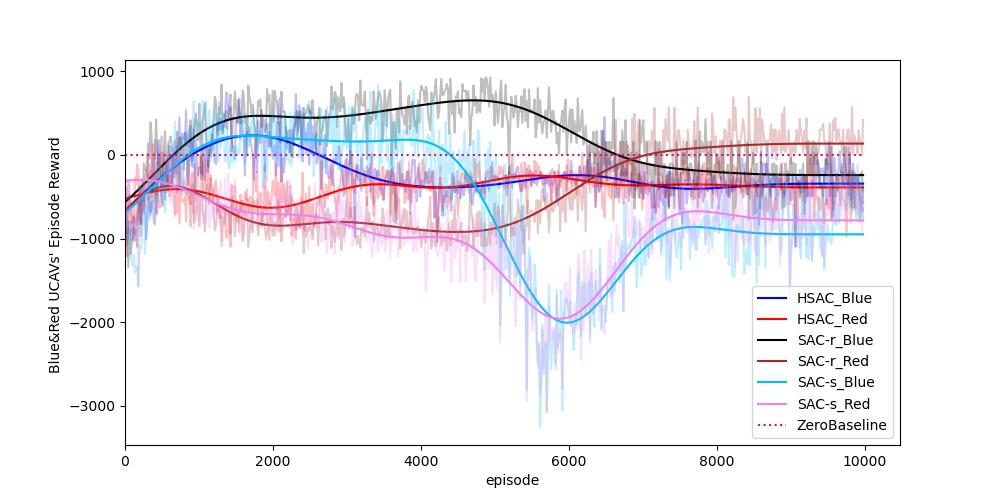}
\caption{The original episode reward of  different methods of $\text{UCAV}^b$ and $\text{UCAV}^r$ in the self-play training process}
\label{Fig:selfTraining}
 \end{subfigure}\\
 \begin{subfigure}[t]{8cm}
\includegraphics[width = 3.8in]{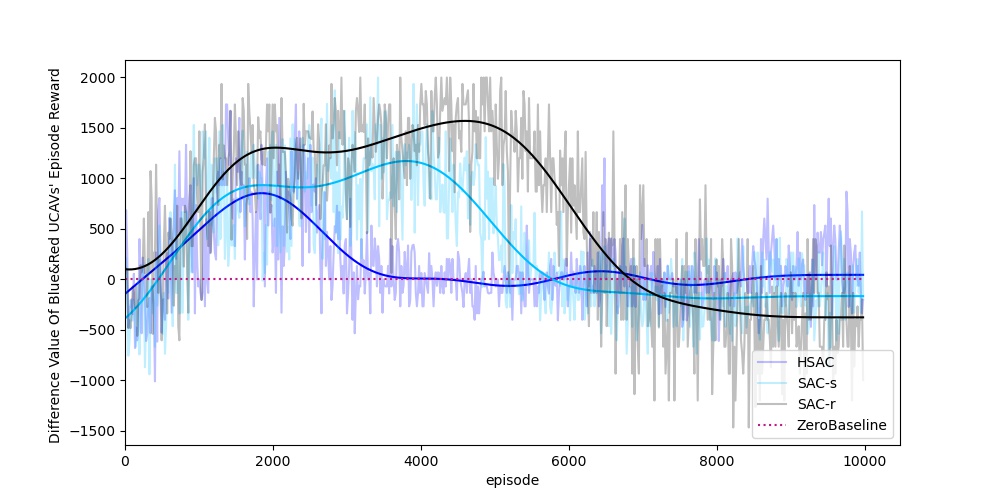}
\caption{Difference values of the episode reward between $\text{UCAV}^b$ and $\text{UCAV}^r$, where the agents of $\text{UCAV}^b$ and $\text{UCAV}^r$ were trained by SAC-s, HSAC, and SAC-r}
\label{Fig:DifferenceValueOfBlue&RedEpisodeReward}
\end{subfigure}\\
\caption{The self-play training process of SAC-s, HSAC, and SAC-r in confrontation task}
\label{Fig:TrainProcess}
\end{figure}

        The training processes of SAC-s, SAC-r, and HSAC are shown in Figure~\ref{Fig:TrainProcess}, after $10^4$ training episodes. 
        Episode reward's variation of $\text{UCAV}^b$ and $\text{UCAV}^r$, in the self-play training process, are shown in Figure~\ref{Fig:selfTraining}.
        Figure~\ref{Fig:DifferenceValueOfBlue&RedEpisodeReward} reflects the difference value of episode reward between $\text{UCAV}^b$ and $\text{UCAV}^r$ during the self-play training process.
   Meanwhile, we record the episode number required for the convergence of the methods, 
   the convergent episode reward of $\text{UCAV}^b$ and $\text{UCAV}^r$, and the difference value between two sides' episode reward in Table~\ref{Tb:SPConverge}.
   We use the same way, mentioned in Algorithm~\ref{algo:HSAC}, to determine whether the methods are converged.
   
        Combining the information in Figure~\ref{Fig:TrainProcess} and Table~\ref{Tb:SPConverge}, we can find that, in the self-play training process, the agents of $\text{UCAV}^b$ and $\text{UCAV}^r$ trained by HSAC can converge to an equilibrium point\cite{nash1951non} in the policy space.
        And the convergence speed of HSAC is the fastest in all three methods.
        Because each of these two sides' UCAVs cannot get much higher episode rewards in the self-play training process, meanwhile, both of the two sides' episode rewards tend to be the same.
        Although the agents trained by SAC-r get the highest scores in this task, as shown in Figure~\ref{Fig:selfTraining}, the policy is not the equilibrium point that we want to find in the policy space, for the reason that this policy can not guarantee the episode reward of $\text{UCAV}^r$ gradually equalling to the episode reward of $\text{UCAV}^b$ during the self-play training process, as shown in Figure~\ref{Fig:DifferenceValueOfBlue&RedEpisodeReward}.
        On the other side, the agent trained by SAC-s cannot get a higher episode reward than the agent trained by HSAC, as illustrated in Figure~\ref{Fig:selfTraining}, although its episode reward difference between two sides' agents can also converge near zero.
        We can find that using HSAC combined with the idea of self-play can quickly converge to an equilibrium point in policy space and the convergent policy performs better than the policy trained via SAC-s.

        After training, the conclusion we can find that the ability of HSAC to find an equilibrium point in the policy space is the best one in these three methods. 
        And the simulation results below can also support this perspective.
        
        \subsubsection{Simulation Results}

        Here, we will evaluate the superiority of HSAC according to the results of the fair 1 vs 1 confrontation for the three agents trained by SAC-s, SAC-r, and HSAC method.
        After $10^5$ confrontations between any two of three agents, we outline the confrontations win rate in Table~\ref{Tb:rwinrate}, where DN denotes the number of draws.
        \begin{table}[hbt]
        \centering
        \scalebox{1.2}{
        \begin{tabular}{| m{1.2cm}<{\centering}m{1cm}<{\centering} | m{1cm}<{\centering} | m{1.8cm}<{\centering} |m{1.3cm}<{\centering}|} 
        \hline
        \diagbox {Blue}{Red}&& SAC-s & SAC-r & HSAC \\ \hline
        \multicolumn{1}{|l|}{SAC-s} & Blue Win &  $34.9\% $& $42.1 \%$ &  $18.7\%$\\ \cline{2-5} 
        \multicolumn{1}{|l|}{}    & Red Win &  $\bm{65.1\%}$  & $ \bm{57.9\%}$ & $ \bm{81.3 \%}$\\ \cline{2-5} 
        \multicolumn{1}{|l|}{}    & DN &  43785  & 372 &  49939\\ \hline
        
        \multicolumn{1}{|l|}{SAC-r} & Blue Win& $\bm{61.0\%}$   & $ 30.5\% $&  $39.4\%$ \\ \cline{2-5} 
        \multicolumn{1}{|l|}{}    & Red Win & $39.0\%$& $\bm{69.5\%}$  & $\bm{60.6\%}$ \\ \cline{2-5} 
        \multicolumn{1}{|l|}{}    & DN &  328  & 1 &  619\\ \hline

        \multicolumn{1}{|l|}{HSAC} & Blue Win &  $\bm{66.8\%}$& $\bm{60.8\%}$  & $45.4\%$  \\ \cline{2-5} 
        \multicolumn{1}{|l|}{}    & Red Win &  $33.2\%$ & $39.1\%$  & $\bm{54.6\%} $ \\ \cline{2-5} 
        \multicolumn{1}{|l|}{}    & DN &  35448  & 843 &  75228\\ \hline
        \end{tabular}
        }
        \caption{The relative win rate of different methods after the agents are evaluated $10^5$ episodes in confrontation task, where DN means the number of draws in $10^5$ episodes}
        \label{Tb:rwinrate}
        \end{table}
        
        The conclusion can be deduced that the agent trained by HSAC is more robust than the other two agents which were trained by SAC-s and SAC-r, from Table~\ref{Tb:rwinrate}.
        We can find that only the win rates of $\text{UCAV}^b$ and $\text{UCAV}^r$ trained by HSAC are similar in the self confrontations and the agent trained by HSAC has more probability to win the confrontation when facing the agents trained by other methods.
        This phenomenon means that the agent trained by HSAC can achieve the best performance whatever the initial state.
        Meanwhile, through Table~\ref{Tb:rwinrate} we can find that the performance of the policies trained by SAC-s and SAC-r are influenced easily by the initial state because the same policy performs really differently when it is used in different sides' $\text{UCAV}$. 
        We also find that using artificial prior experience, such as SAC-r, may have strong directivity and the trained agent can hardly achieve the draw situation in the confrontation game.
        This phenomenon means this method may not be an applicable way to train the agent in a self-play confrontation task, which may result in the deviation of the original task's goal.

\begin{figure}[hbt]
        \centering
         \scalebox{1}{
        \includegraphics[width = 3in]{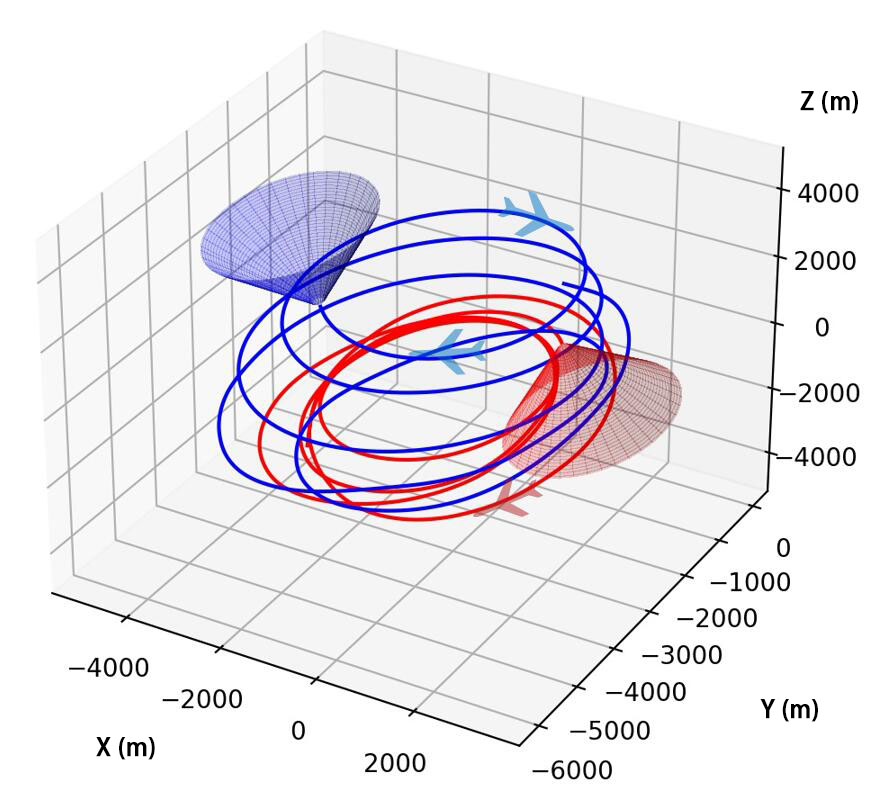}
            }
        \caption{The process of the agent trained by HSAC combat with the agent trained by SAC-s, the agent trained by HSAC play the role of $\text{UCAV}^b$ and the agent trained by SAC-s play the role of $\text{UCAV}^r$}
        \label{Fig:hVSS}
         \end{figure}

\begin{figure}[hbt]
        \centering
         \scalebox{1}{
        \includegraphics[width = 3in]{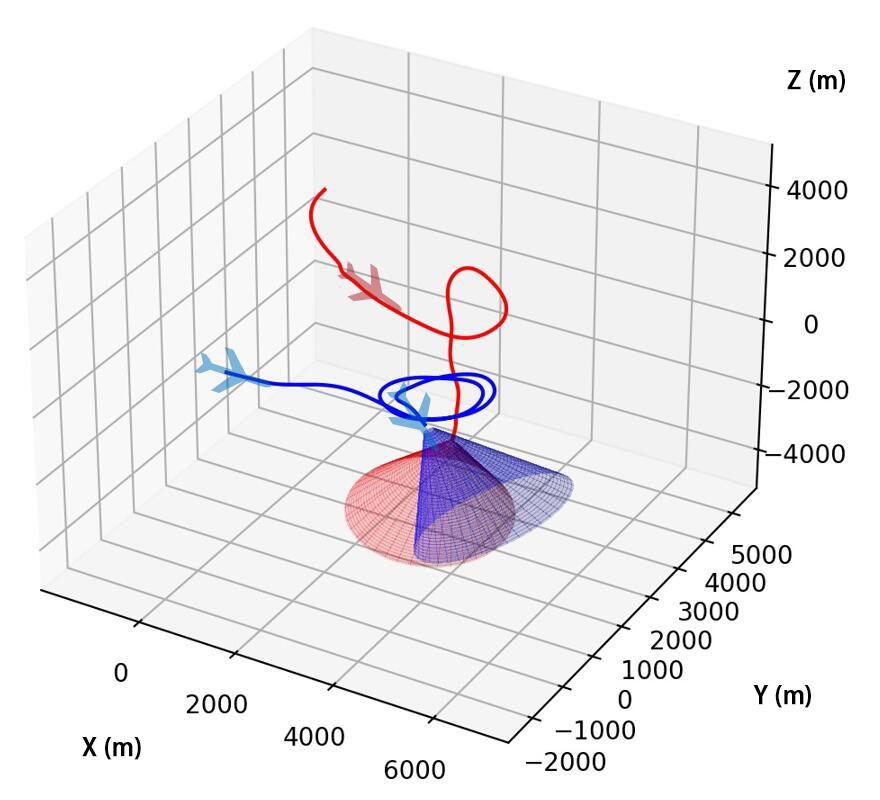}
            }
        \caption{The process of the agent trained by HSAC combat with the agent trained by SAC-r, the agent trained by HSAC play the role of  $\text{UCAV}^b$ and the agent trained by SAC-r play the role of  $\text{UCAV}^r$}
        \label{Fig:hVSRC}
         \end{figure}

\begin{figure}[hbt]
        \centering
         \scalebox{1}{
        \includegraphics[width = 3in]{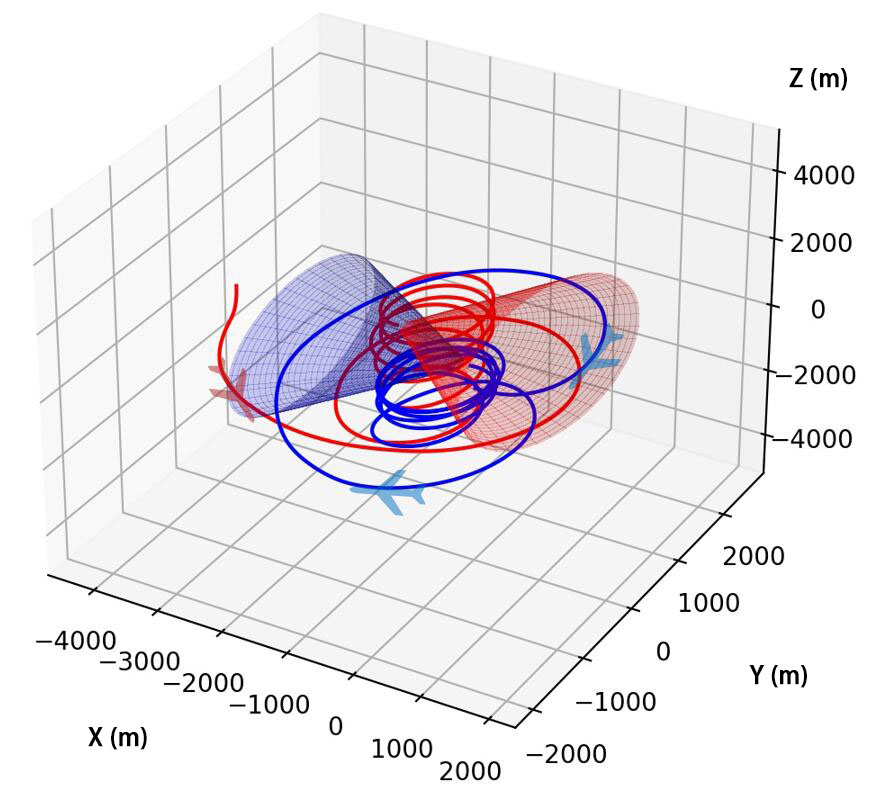}
            }
        \caption{The process of the agent trained by HSAC combat with an agent with the same policy}
        \label{Fig:hVSH}
         \end{figure}

        To illustrate the performance of HSAC, intuitively, we will demonstrate the attack and defense situation in the confrontation task. The processes of the agent trained by HSAC combat with the agent trained by SAC-s, SAC-r, and HSAC are shown in Figure~\ref{Fig:hVSS},  Figure~\ref{Fig:hVSRC},  Figure~\ref{Fig:hVSH}, respectively. 
       
        As shown in Figure~\ref{Fig:hVSS} and Figure~\ref{Fig:hVSH}, the agents trained by the SAC-s method and HSAC method are falling into an equilibrated situation.
        Also, it is obvious that the pursuing and escaping ability of the agent trained by HSAC seems better than the agent trained by SAC-s, from these two figures combined with the information in Table~\ref{Tb:rwinrate}.
        And the agent trained by SAC-r performs unsatisfactorily when combating with the agent trained by HSAC, as shown in Figure~\ref{Fig:hVSRC}. This means that SAC-r may bias the optimal converging direction of the original problem.

        In conclusion, SAC-r is unsuited in the confrontation task, such as air combat.
        SAC-s may spend too much time exploring the suitable policy.
        However, through HSAC we proposed, the artificial prior experience used in SAC-r can be a guide to the agent at the beginning and would not change the optimal converging direction of the original problem.
        With the extra reward $R^{extra}$ being ablated along with the training process gradually, HSAC can solve the sparse reward RL problem in air combat scenarios without the bias in the optimal converging direction. 

    
\section{Conclusion}\label{SE:Conclusion}

In this paper, an elaborate 3D air combat simulation environment is constructed for the RL-based methods training, where the state space is compressed through the relative perspective of the UCAVs, and the initial state is designed to mimic radar detection.
All of those designs narrow the gap between the simulation and the reality, which
can help this air combat policy trained in the simulation environment migrate to reality.

A new method to solve the sparse reward RL problem called the HSAC method is proposed in this paper, which can well balance the convergence rate speed up and the optimal solution deviation resulting from artificial priors intervention.
The proofs of the convergence and the feasibility of HSAC are also given in this paper.
During the HSAC training process, the RL agent can be guided to a feasible policy at the beginning and then gradually forget the artificial prior experience reward to avoid biasing the goal of the original task.
It has been proved by the experiments that this method is really effective in the severe sparse reward RL problem of the air combat scenarios.

After evaluating the training process of HSAC, SAC-s, and SAC-r, we find that HSAC can get the highest episode reward in attack horizontal flight UCAV task. Meanwhile, the HSAC method can find the desired nash equilibrium policy greatly faster than another two methods in the self-play training process, and this policy has a great capability of both offense and defense.
In future works, we will extend our method to the multi-agent air combat scenarios and we hope the agent can gain the ability of both competition and cooperation.

 \section*{}

\bibliographystyle{unsrtnat}
\bibliography{references}
\newpage
\appendix

\section{Proof of the equation $F_3(\cdot)$}\label{Appendix:A}
The equation of $F_3(\cdot)$ can be expanded as follow:
\begin{align}
\centering
    \begin{split}
        F_3(\phi,q)=&(1-1)F_2(\phi)+qF_1(\phi)\\
                =& (1-q)\mathbb{E}_{s_0\sim {\rho _0}({s_0})}
        \left[{\sum\limits_{t = 0}^T {{\gamma^t} \big[  {R}({s_t},{a_t})} +{R^{extra}}({s_t},{a_t}) \big]} \right] + q \mathbb{E}_{s_0\sim {\rho _0}({s_0})}
        \left[{\sum\limits_{t = 0}^T {{\gamma^t} \big[  {R}({s_t},{a_t})}) \big]} \right] \\
        =& \mathbb{E}_{s_0\sim {\rho _0}({s_0})}
        \left[(1-q){\sum\limits_{t = 0}^T {{\gamma^t} \big[  {R}({s_t},{a_t})} +{R^{extra}}({s_t},{a_t})\big]
        + q \sum\limits_{t = 0}^T {{\gamma^t} \big[  {R}({s_t},{a_t})} 
        \big]} \right]\\
        =& \mathbb{E}_{s_0\sim {\rho _0}({s_0})}
        \left[{\sum\limits_{t = 0}^T {\gamma^t} 
        \big[  
        (1-q)\big({R}({s_t},{a_t}) +{R^{extra}}({s_t},{a_t})\big)
        + q   {R}({s_t},{a_t})} 
        \big]
         \right]\\
        = & \mathbb{E}_{s_0\sim {\rho _0}({s_0})}
        \left[{\sum\limits_{t = 0}^T {\gamma^t} \big[  R^{\text{H}}({s_t},{a_t},q) \big]} \right]\\
    \end{split}
\end{align}
The homotopy NLP problem mentioned in \textbf{NLP3 } can be described as follow:
\begin{align}
   \centering
    \begin{split}
        \mathop {\max }\limits_{\phi} F_3(\phi,q) = & 
        \mathbb{E}_{s_0\sim {\rho _0}({s_0})}
        \left[{\sum\limits_{t = 0}^T {{\gamma^t} \big[  {R^\text{H}}({s_t},{a_t},q)}  \big]} \right]\\
        h_j(\phi)=&0 \qquad j=1,\cdot\cdot\cdot,m\\
        g_j(\phi)\ge&0 \qquad j=1,\cdot\cdot\cdot,s
    \end{split}
    \label{Eq:A1}
    \end{align}
    
\section{Homotopy Path Existence and Convergence of the Path Following Method} \label{Appendix:B}
\subsection{Formalize the Problem as $H(\cdot)=0$ }
The $s$ in the NLP problem of $F_3(\cdot)$, in ~\ref{Appendix:A}, can also be described as state transition function $T(\cdot)$:
  \begin{align}
  \begin{split}
    F_3(\phi,q)=\mathbb{E}_{s_0\sim {\rho _0}({s_0})}
    \left[
    \sum\limits_{t = 0}^\mathrm{T} {\gamma^{t}
    \big[R^{{\rm{Homotopy}}}\left(
    T\left(s^{t},\pi_{\phi}(s^{t})\right),q
    \right)
    \big]}
    \right]
      \end{split}
 \end{align}
which means that the function $F_3(\cdot)$ only relates to parameter $\phi$ and $q$.
Then using the Karush-Kuhn-Tucker (KKT) equations \cite{fiacco1990nonlinear} to formalize the problem described in Eq.~(\ref{Eq:A1}) as Eq.~(\ref{Eq:H()})
    \begin{align}
        H(\phi ,\alpha ,\mu ,q) = \left[ {\begin{array}{*{20}{c}}
        {{\nabla _\phi }f(\phi ,q) + \sum\limits_{j = 1}^s {\alpha _j^ + {\nabla _\phi }{g_j}(\phi ) + } \sum\limits_{j = 1}^m {{\mu _j}{\nabla _\phi }{h_j}(\phi )} }\\
        {\alpha _1^ -  - {g_1}(\phi )}\\
        \cdot \\
        \cdot \\
        \cdot \\
        {\alpha _s^ -  - {g_s}(\phi )}\\
        {{h_1}(\phi )}\\
        \cdot \\
        \cdot \\
        \cdot \\
        {{h_m}(\phi )}
        \end{array}} \right]
        \label{Eq:H()}
    \end{align}
     Here, $H(\cdot)$ is called the homotopy function. And
     $H:\mathbb{R}^{n+m+s+1}\to\mathbb{R}^{n+m+s}$ because $\phi \in \mathbb{R}^n$,$\alpha \in \mathbb{R}^s$,$\mu \in \mathbb{R}^m$, and $q\in\mathbb{R}^1$. We let $\alpha^{+}_{j}=\left[max\left\{0,\alpha_j\right\}\right]^3$ and $\alpha^{-}_{j}=\left[max\left\{0,-\alpha_j\right\}\right]^3$, $\mu$ here are only the auxiliary variables.
    The problem mentioned in \textbf{NLP3} equals to solve the KKT equation $H(\phi ,\alpha ,\mu ,q)=0$. 
    
\subsection{Hmotopy Path Existence}\label{existence}
The idea of the homotopy method is following a path to a solution, where the "path" meant a piecewise differentiable curve in solution space. 

Zangwill\cite{lemke1984pathways} define the set of all solutions as $H^{-1}$.
\begin{definition}
    Given a homotopy function $H:\mathbb{R}^{n+1}\to \mathbb{R}^{n}$, we must now be more explicit about solutions to 
        \begin{align}
            H(\phi ,\alpha ,\mu,q)=0
        \end{align}
    In particular, define 
        \begin{align}
             H^{-1}={\left\{(\phi ,\alpha ,\mu,q)\big| H(\phi ,\alpha ,\mu,q)=0 \right\}}
        \end{align}
    as the set of all solutions $(\phi ,\alpha ,\mu,q)\in \mathbb{R}^{n+1}$
\end{definition}

With the implicit function theorem can ensure that $H^{-1}$ consists solely of paths\cite{lemke1984pathways}. The Jacobian of homotopy function $H(\phi ,\alpha ,\mu,q)$ can be written as an $(n+m+s)\times (n+m+s+1)$ matrix as shown in Eq.~(\ref{Eq:HJMatrix}).

Then the existence of the path was given in \cite{lemke1984pathways}:
\begin{theorem}[Path Existence]\cite{lemke1984pathways}
Let $H:\mathbb{R}^{n+m+s+1}\to\mathbb{R}^{n+m+s}$ be continuously differentiable and suppose that for every $(\phi ,\alpha ,\mu,q)\in H^{-1}$, the Jacobian $H'(\phi ,\alpha ,\mu,q)$ is of full rank. Then $H^{-1}$ consists only of continuously differentiable paths.
\label{th:PathExist}
\end{theorem}


%

\begin{align}
\centering
\begin{split}
     H'(\phi ,\alpha ,\mu,q) = 
     \left( {\begin{array}{*{20}{c}}
    {\frac{{\partial {H_1}}}{{\partial {\phi_1}}}}
    &{ \cdot  \cdot  }
    & {\frac{{\partial {H_1}}}{{\partial {\phi_n}}}}
    & {\frac{{\partial {H_1}}}{{\partial {\alpha_1}}}}
    &{ \cdot  \cdot   }
    & {\frac{{\partial {H_1}}}{{\partial {\alpha_s}}}}
     & {\frac{{\partial {H_1}}}{{\partial {\mu_1}}}}
        &{ \cdot  \cdot   }
    & {\frac{{\partial {H_1}}}{{\partial {\mu_m}}}}
    &{\frac{{\partial {H_1}}}{{\partial q}}}\\
    \cdot &{}& \cdot & \cdot& {}
    &\cdot& \cdot &{}& \cdot & \cdot\\
\cdot &{}& \cdot & \cdot& {}
    &\cdot& \cdot &{}& \cdot & \cdot\\
    {\frac{{\partial {H_{n+m+s}}}}{{\partial {\phi_1}}}}
    &{ \cdot  \cdot  }
    & {\frac{{\partial {H_{n+m+s}}}}{{\partial {\phi_n}}}}
    & {\frac{{\partial {H_{n+m+s}}}}{{\partial {\alpha_1}}}}
    &{ \cdot  \cdot   }
    & {\frac{{\partial {H_{n+m+s}}}}{{\partial {\alpha_s}}}}
     & {\frac{{\partial {H_{n+m+s}}}}{{\partial {\mu_1}}}}
    &{ \cdot  \cdot   }
    & {\frac{{\partial {H_{n+m+s}}}}{{\partial {\mu_m}}}}
    &{\frac{{\partial {H_{n+m+s}}}}{{\partial q}}}
    \end{array}} \right)
    \label{Eq:HJMatrix}
\end{split}
\end{align}

As the other scholars do in homotopy optimization\cite{lemke1984pathways}, we give an assumption of the Jacobian matrix $H^{'}(\cdot)$:
    \begin{assumption}
    $H'(\phi,\alpha,\mu,q)$ is of full rank for all $(\phi,\alpha,\mu,q)\in H^{-1}$. 
    \label{asp:fullRank}
    \end{assumption}
Combining the assumption~\ref{asp:fullRank} and Theorem~\ref{th:PathExist}, the existence of the homotopy path in this problem can be assured.

\subsection{Convergence of Path Following Method}

The homotopy path existence has been proved in Section ~\ref{existence}. An then we will give the proof of the convergence and the feasibility of the path following method.
In this paper, the corrector-predictor method, with horizontal corrector and elevator predictor, is used to follow the homotopy path along with the iterations. 
To ensure the convergence and the feasibility of this corrector-predictor path-following method, we quote the Theorem ~\ref{thm:PCM} proposed by Zangwill \cite{lemke1984pathways}.
    
    \begin{theorem}[Method Convergence]
    For a homotopy $H\in C^2$ at any $y\in H^{-1}$, let $H'_{\phi}(\phi,q)$ be of full rank. Also, for $(\phi_0,q_0)\in H^{-1}$, let the path be of finite length. Now suppose that a predictor-corrector method uses the horizontal corrector and the elevator predictor.\\
    Given $\varepsilon_1>\varepsilon_2>0 $ sufficiently small, if for all iterations $k$ the predictor step length $d^k$ is sufficiently small, then the method will follow the entire path length to any degree of accuracy.
    \label{thm:PCM}
    \end{theorem}
Where, in traditional predictor-corrector method\cite{lemke1984pathways},  Eq.~(\ref{Eq:P}) and Eq.~(\ref{Eq:C}) are used to decide if the solution satisfy the accuracy requirement in the predictor and corrector step.
\begin{align}
\left\| H(\phi_n,\alpha_n,\mu_n,q_n) \right\|<\varepsilon_1 
\label{Eq:P}
\\
\left\|
 H(\phi_{n+1},\alpha_{n+1},\mu_{n+1},q_{n+1}) \right\|<\varepsilon_2
 \label{Eq:C}
\end{align}
In this paper we suppose $\varepsilon_1 =\varepsilon_2 =\varepsilon$ to be convenient to calculate.
    
    Through the Theorem~\ref{thm:PCM}, we can guarantee the convergence and the feasibility of the original problem mentioned in \textbf{NLP1} with the help of the function $F_3(\cdot)$, using the predictor-corrector path-following method. 

\section{Paramters}
The parameter of the dynamics of UCAV and the criterion of air combat scenario mentioned in Section~\ref{SE:Air_combat_model} are given in Table~\ref{tb:aircombatpar}.
\begin{table}[H]
\centering
\scalebox{1.1}{
\begin{tabular}{l|l}
\hline
Single UCAV Parameter & Value \\ \hline
 mass of UCAV & 150 $kg$\\
 max load factor ($n_{max})$& 10 $g$\\
 velocity band & $[80,400]$ $m/s$\\
 height range & $[2000,8000]$ $m$\\
 max thrust ($T_{max})$&100 $kg$\\
 range of $\dot{\alpha}$&$[-5,5]$ $\deg/s$\\
 range of $\dot{\mu}$&$[-50,50]$ $\deg/s$\\
  range of ${\alpha}$&$[-15,15]$ $\deg$\\\hline
 Combat Parameter & Value\\\hline
 optimum attack range$(d_{min},d_{max})$& (200,3000) $m$\\
 max iteration step & 2000\\
\hline
\end{tabular}
}
 \caption{Air Combatt Simulation Parameters}
\label{tb:aircombatpar}
\end{table}

The hyperparameters of the HSAC algorithm, we mentioned in Section~\ref{SE:HSAC}, are given in Table~\ref{tb:HSACParameters}.
\begin{table}[H]
\centering
\scalebox{1.1}{
\begin{tabular}{l|l}
\hline
Parameter & Value \\ \hline
 optimizer& Adam \\
learning rate & $3\cdot10^{-4}$ \\
discount ($\gamma$)&0.996  \\
number of hidden layers(all networks) & 3  \\
number of hidden units per layer & 256 \\
number of samples per minibatch &  256\\
nonlinearity & ReLU \\ 
replay buffer size&$10^6$\\
entropy target& -dim($A$)\\
target smoothing coefficient ($\tau$)&0.005\\
policy gradient buffer ($M$)&$10^4$\\
total number of homotopy iteration ($N$)&100\\
threshold of $k_{\nabla \pi}$ ($\varepsilon$)&$10^{-5}$\\
\hline
\end{tabular}
}
 \caption{HSAC Hyperparameters}
\label{tb:HSACParameters}
\end{table}

In Table~\ref{tb:4initialS}, we give the four typical initial states which are used to evaluate the quality of the policy trained by different methods in attack horizontal flight UCAV task.
\begin{table}[H]
\centering
\scalebox{1.1}{
\begin{tabular}{|ll|l|l|l|l|l|l|}
\hline
Initial state  &  &$X_{m}$  & $Y_{m}$  &  $H_{m}$  & $V_{m/s}$ & $\chi_{^\circ}$   \\ \hline
\multicolumn{1}{|l|}{Advantageous} & Blue & 0 &0  & 5000 &150 & 45 \\ \cline{2-7} 
\multicolumn{1}{|l|}{} &Red  & 5000 & 5000  & 5000 & 150  & 45 \\ \hline
\multicolumn{1}{|l|}{Disadvantageous} &Blue & 0 &0  & 5000 &150  & -45 \\ \cline{2-7} 
\multicolumn{1}{|l|}{} & Red &-5000  &5000  & 5000 & 150 &  -45\\ \hline
\multicolumn{1}{|l|}{Head-On} & Blue & 0 &0  & 5000 &150  & 45 \\ \cline{2-7} 
\multicolumn{1}{|l|}{} & Red& 5000 & 5000  & 5000 & 150 & -135  \\ \hline
\multicolumn{1}{|l|}{Neutral} & Blue & 0 &0  & 5000 &150  & 45 \\ \cline{2-7} 
\multicolumn{1}{|l|}{} & Red & 5000 & -5000  & 5000 & 150  & -135 \\ \hline
\end{tabular}
}
\caption{Initial state settings for evaluating the quality of agents }
\label{tb:4initialS}
\end{table}

\section{Performance of Different Methods in Attack Horizontal Flight UCAV Task}
\begin{figure*}[hbt]
            \centering
                \begin{subfigure}[t]{4.5cm}
                \centering
                \includegraphics[width = 4.5cm]{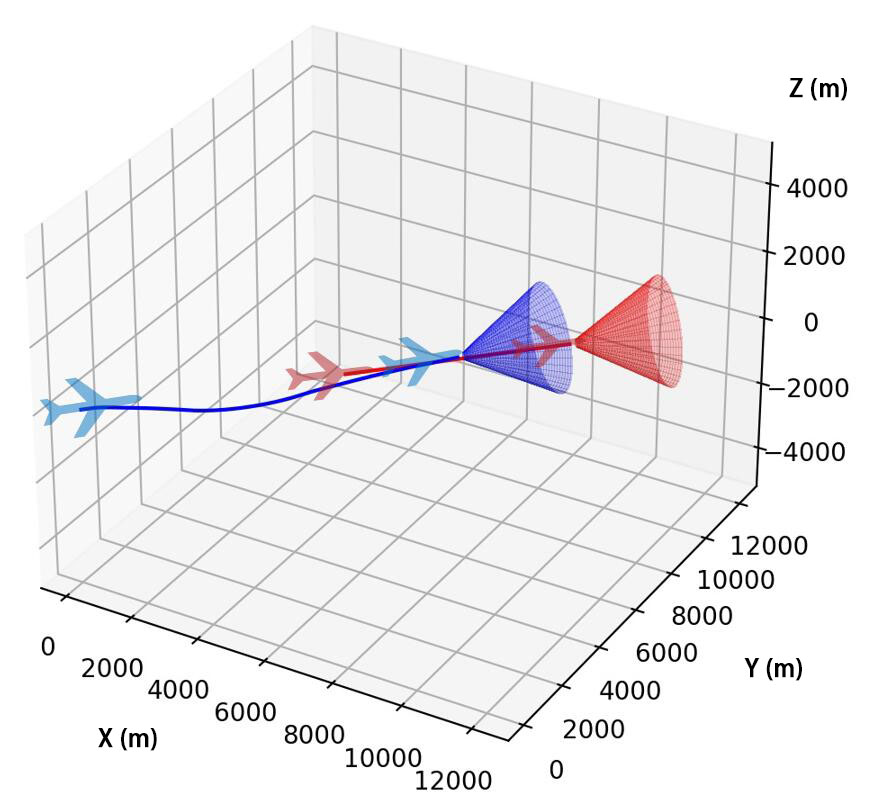}
                \caption{Agent trained by SAC-s in an advantageous situation}
                \label{Fig:AttackLevelFly_S_advantage}
                \end{subfigure}
                ~~~~~
                \begin{subfigure}[t]{4.5cm}
                \centering
                \includegraphics[width = 4.5cm]{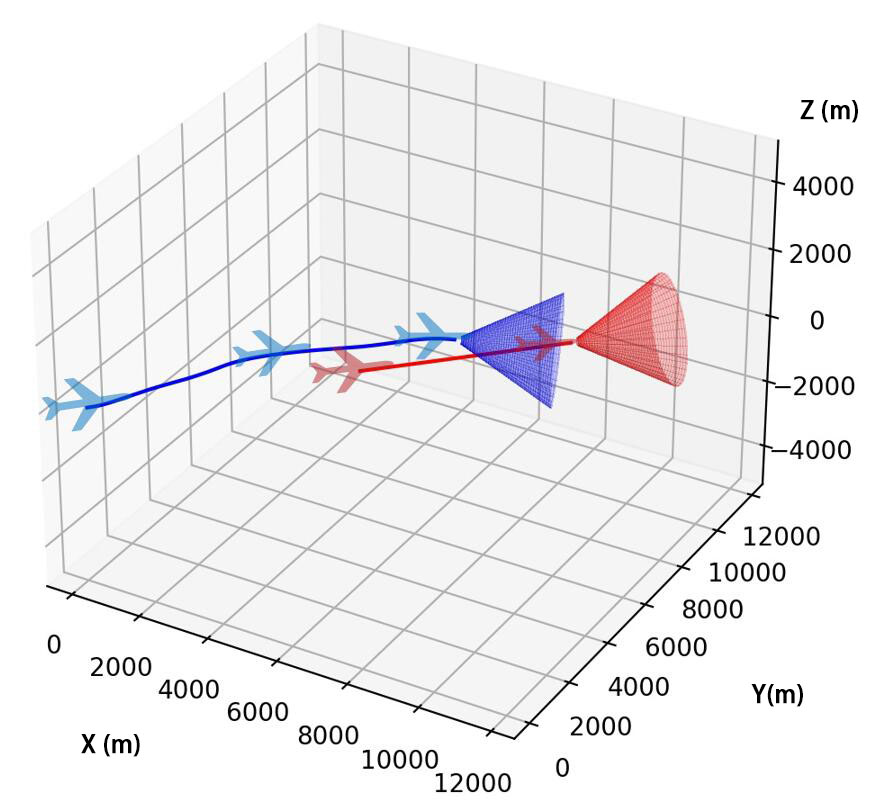}
              
                \caption{Agent trained by SAC-r in an advantageous situation}
                \label{Fig:AttackLevelFly_RC_advantage}
                \end{subfigure}
                ~~~~~
                \begin{subfigure}[t]{4.5cm}
                \centering
                \includegraphics[width = 4.5cm]{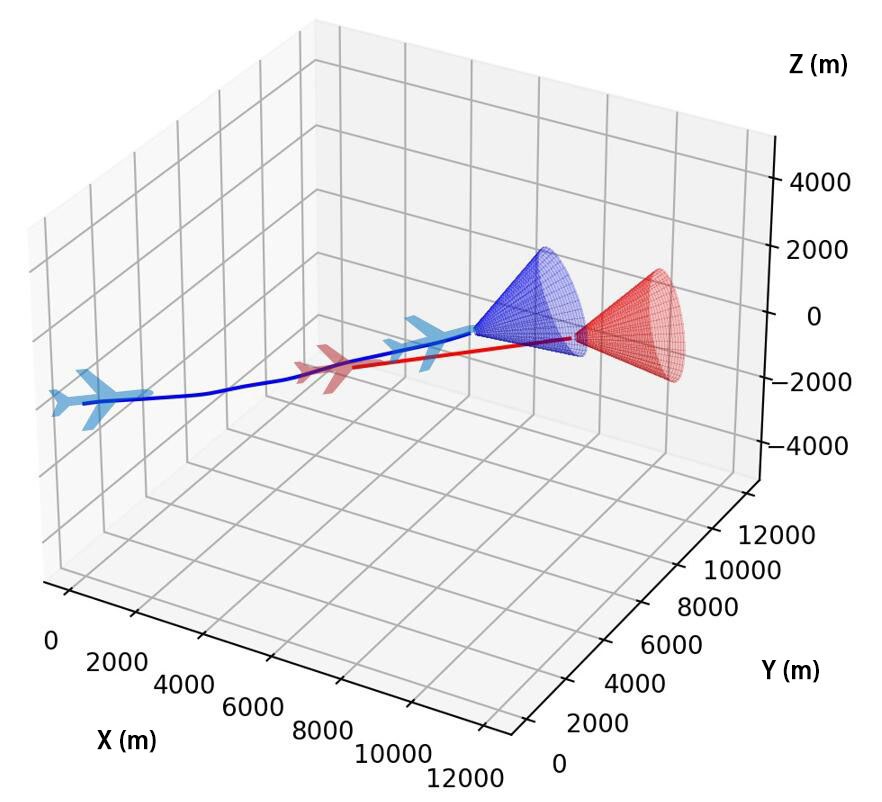}
                \caption{Agent trained by HSAC in an advantageous situation}
                \label{Fig:AttackLevelFly_TL_advantage}
                \end{subfigure}\\
                \begin{subfigure}[t]{4.5cm}
                \centering
                \includegraphics[width = 4.5cm]{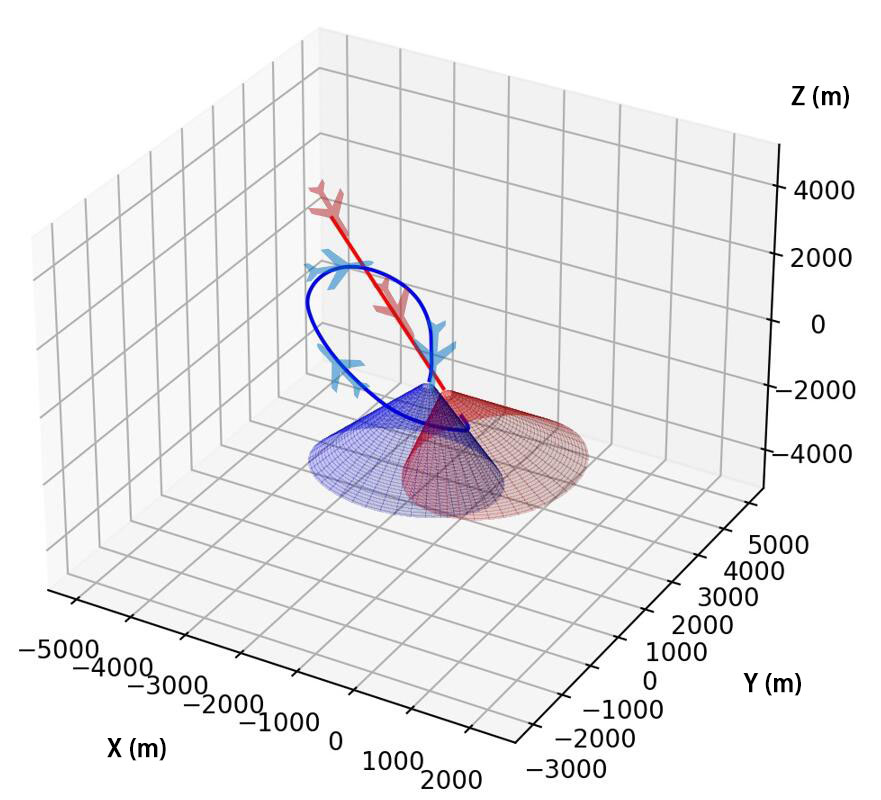}
                \caption{Agent trained by SAC-s in a disadvantageous situation}
                \label{Fig:AttackLevelFly_S_disadvantage}
                \end{subfigure}
                ~~~~~
                \begin{subfigure}[t]{4.5cm}
                \centering
                \includegraphics[width = 4.5cm]{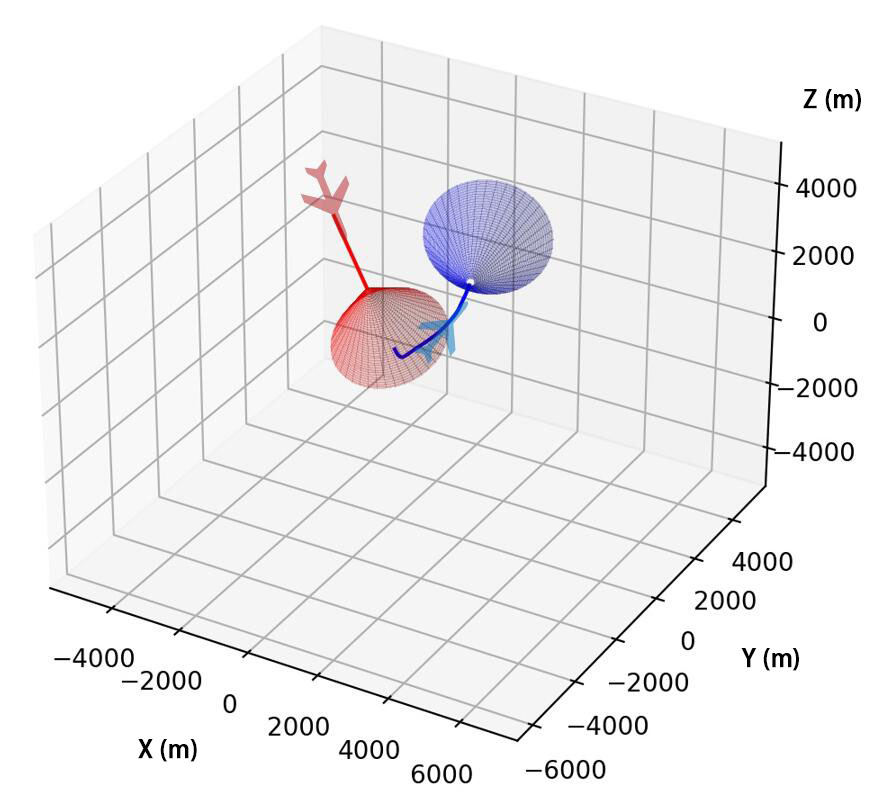}
              
                \caption{Agent trained by SAC-r in a disadvantageous situation}
                \label{Fig:AttackLevelFly_RC_disadvantage}
                \end{subfigure}
                ~~~~~
                \begin{subfigure}[t]{4.5cm}
                \centering
                \includegraphics[width = 4.5cm]{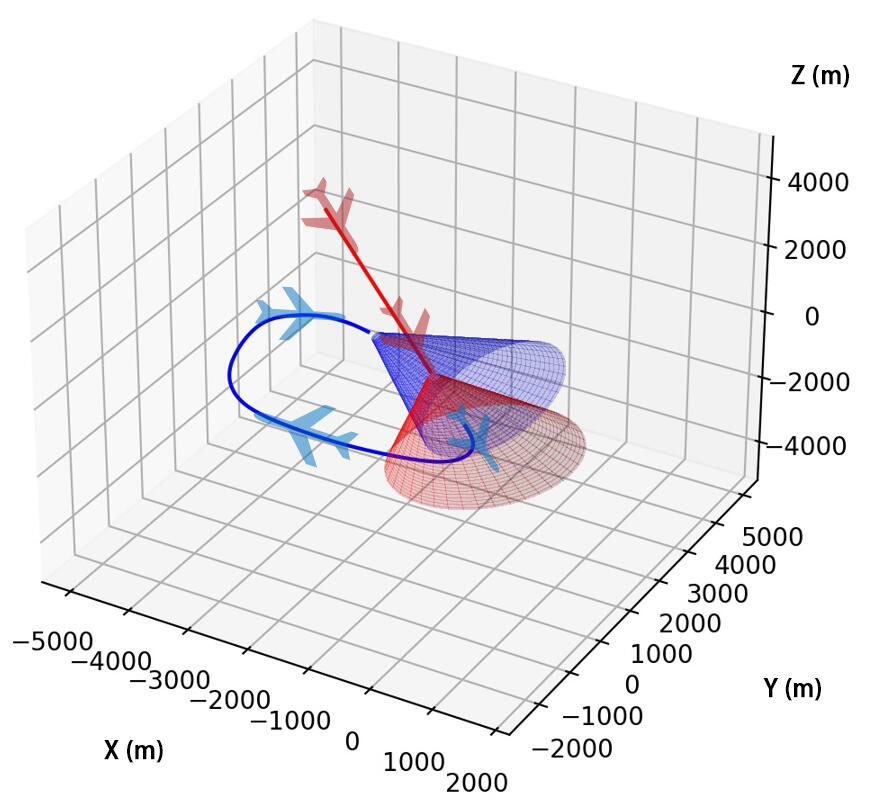}
                \caption{Agent trained by HSAC in a disadvantageous situation}
                \label{Fig:AttackLevelFly_TL_disadvantage}
                \end{subfigure}\\
                \begin{subfigure}[t]{4.5cm}
                \centering
                \includegraphics[width = 4.5cm]{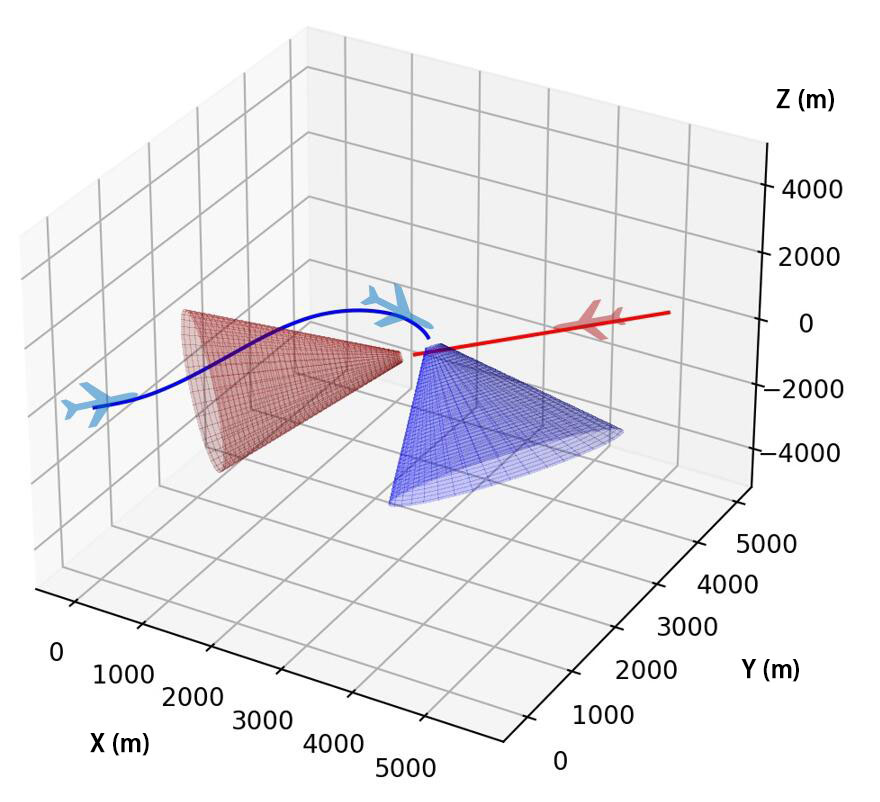}
                \caption{Agent trained by SAC-s in a head-on situation}
                \label{Fig:AttackLevelFly_S_balance}
                \end{subfigure}
                ~~~~~
                \begin{subfigure}[t]{4.5cm}
                \centering
                \includegraphics[width = 4.5cm]{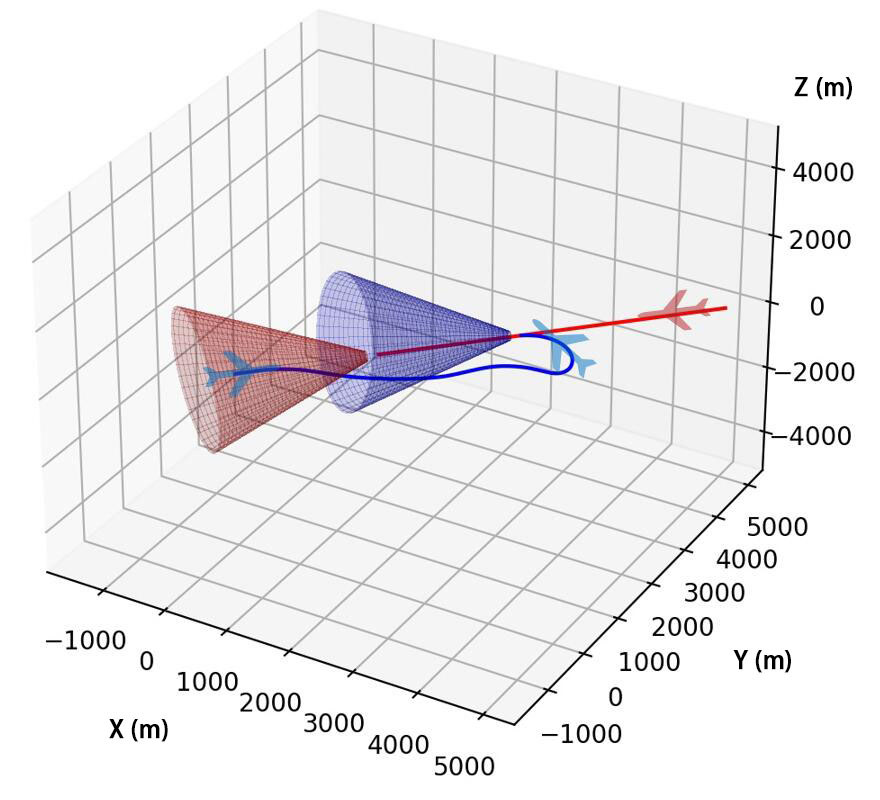}
              
                \caption{Agent trained by SAC-r in a head-on situation}
                \label{Fig:AttackLevelFly_RC_balance}
                \end{subfigure}
                ~~~~~
                \begin{subfigure}[t]{4.5cm}
                \centering
                \includegraphics[width = 4.5cm]{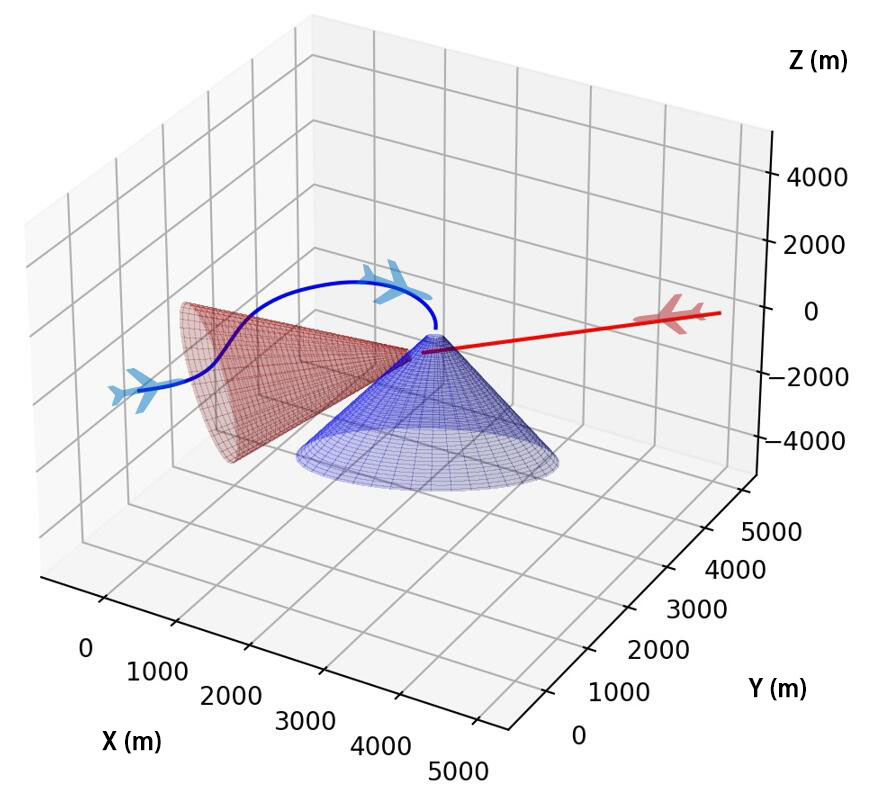}
                \caption{Agent trained by HSAC in a head-on situation}
                \label{Fig:AttackLevelFly_TL_balance}
                \end{subfigure}\\
                \begin{subfigure}[t]{4.5cm}
                \centering
                \includegraphics[width = 4.5cm]{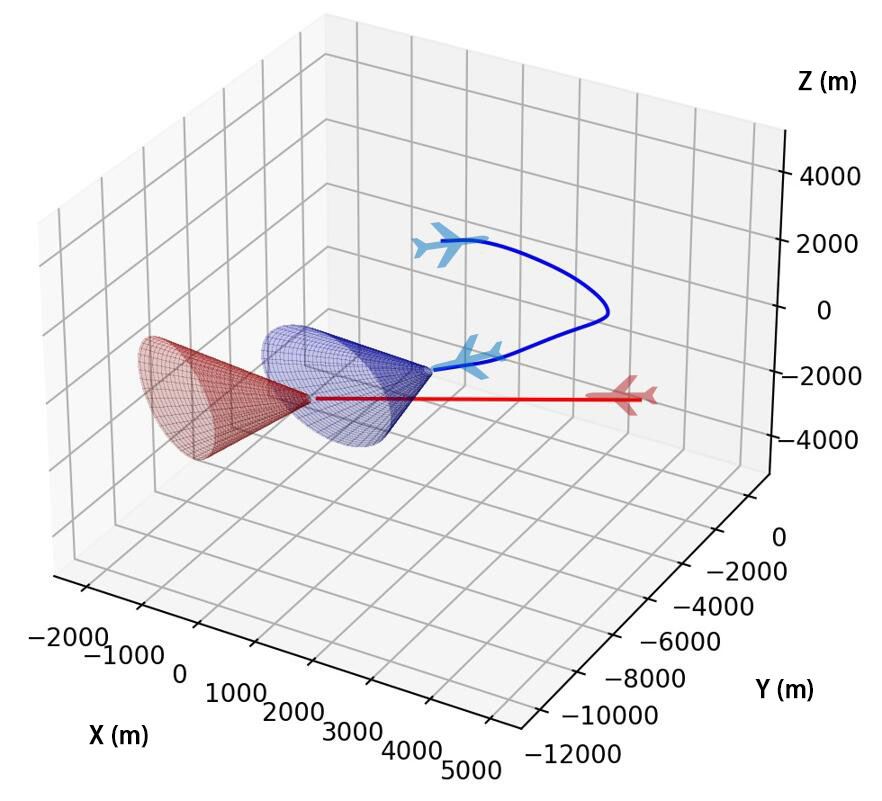}
                \caption{Agent trained by SAC-s in a neutral situation}
                \label{Fig:AttackLevelFly_S_netural}
                \end{subfigure}
                ~~~~~
                \begin{subfigure}[t]{4.5cm}
                \centering
                \includegraphics[width = 4.5cm]{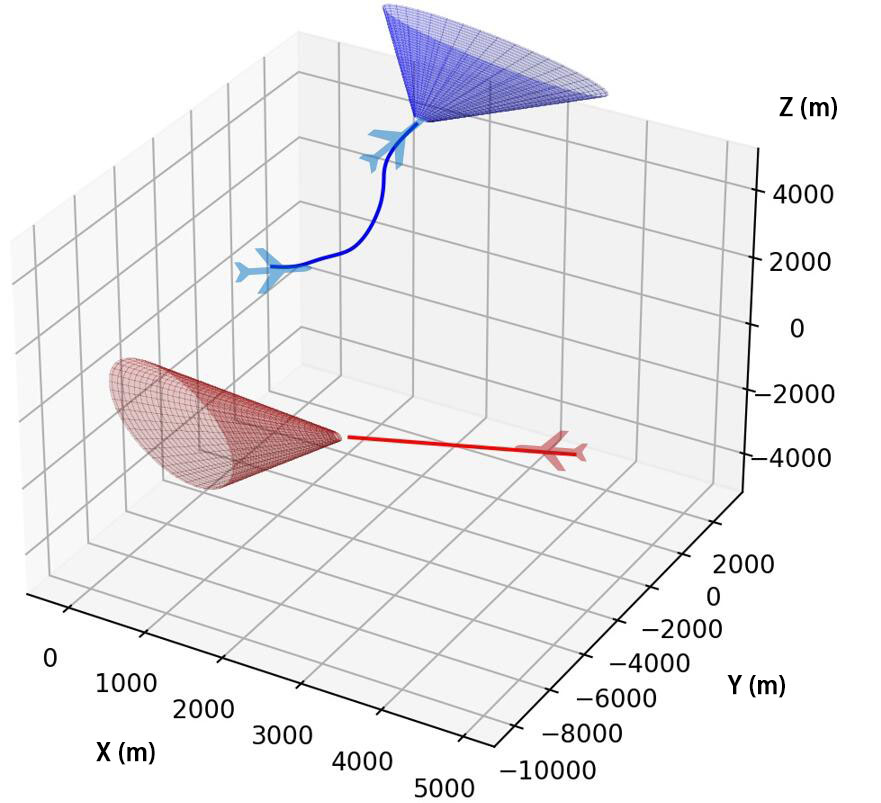}
              
                \caption{Agent trained by SAC-r in a neutral situation}
                \label{Fig:AttackLevelFly_RC_neutral}
                \end{subfigure}
                ~~~~~
                \begin{subfigure}[t]{4.5cm}
                \centering
                \includegraphics[width = 4.5cm]{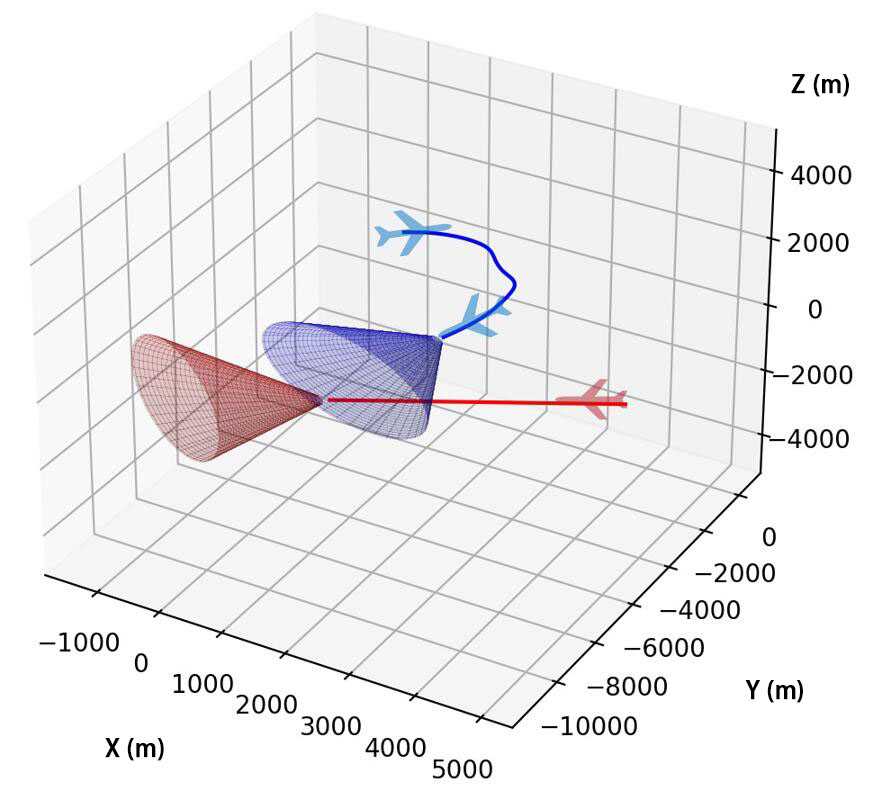}
                \caption{Agent trained by HSAC in a neutral situation}
                \label{Fig:AttackLevelFly_TL_neutral}
                \end{subfigure}
            \caption{Performance of agents trained by different methods with different initial situations in the attack horizontal flight UCAV task }
            \label{Fig:DAIDS}
            \end{figure*}

\end{document}